\pdfoutput=1
\documentclass[11pt]{article}

\usepackage[final]{acl}

\usepackage{times}
\usepackage{latexsym}

\usepackage[T2A,T1]{fontenc}
\usepackage[german,english]{babel}
\babeltags{de=german}

\usepackage{linguex}

\usepackage[utf8]{inputenc}

\usepackage{microtype}

\usepackage{inconsolata}

\usepackage{graphicx}

\usepackage{booktabs}
\usepackage{enumitem}
\usepackage{multirow}
\usepackage{wasysym}  % for slanted arrows

\usepackage{colortbl}
\usepackage{xcolor}

\usepackage{makecell}
\usepackage{array}
\usepackage{caption} % Required for \captionof
\usepackage{subcaption}
\usepackage{relsize}  % for \textsmaller{}

\title{Translationese as a Rational Response to Translation Task Difficulty}
\author{Maria Kunilovskaya \\
	University of Technology Nuremberg\thanks{This work was carried out while the author was affiliated with the Department of Language Science and Technology, Saarland University.}\\ %\\ Department of Computer Science \& Artificial Intelligence, NLLG lab\\
	\texttt{maria.kunilovskaya@utn.de} 
}
\begin{document}
\maketitle

\begin{abstract}
Translated texts exhibit systematic differences from comparable texts originally written in the target language. Explaining this phenomenon, commonly known as \textit{translationese}, remains an open challenge. Translationese has been attributed to production tendencies (e.g. interference, simplification), socio-cultural variables, and language-pair effects, yet a unified explanatory account is lacking. 
We investigate the hypothesis that translationese is a response to the cognitive load inherent in the translation task. We test whether observable translationese can be predicted from quantifiable measures of translation task difficulty. Translationese is measured as a segment-level probability of being a translation produced by an automatic classifier (\textit{translatedness score}). Translation task difficulty includes source-text and cross-lingual transfer components. They are captured by information-theoretic metrics based on LLM surprisal and by established syntactic and semantic alternatives. We use a bidirectional English-German corpus comprising written and spoken subcorpora.
Results indicate that translationese can only be explained in part by translation task difficulty, especially in the English-to-German direction. For most experiments, cross-lingual transfer difficulty contributes more than source-text complexity. Information-theoretic indicators match or outperform traditional features in written mode, but offer no advantage in spoken mode. Source-text syntactic complexity and translation-solution entropy emerged as the strongest predictors of translationese across language pairs and modes.
\end{abstract} 

\section{Introduction}
\label{sect:intro}
Previous work in translation studies yielded abundant evidence that any cross-linguistic mediation exhibits systematic statistical deviations from the patterns observed in comparable non-translated, originally-authored texts in the target language. This property of translations is usually referred to as \textit{translationese}~\cite{Gellerstam1986}.
Evidence from corpus-based and computational studies confirms that translationese is pervasive, reinforcing its view as an intrinsic feature of cross-linguistic mediation, which produces a translational variant of the target language, often described as a ``third code''~\cite{Frawley1984, baker1993corpus}.

Crucially, this phenomenon appears to be largely independent of translation quality. In fact, one could argue that texts entirely devoid of translationese may border on transcreation, adaptation, or similar forms of mediated communication that allow the source message to be substantially modified in response to the needs of the target audience or a different communicative purpose. Such modifications may involve, for example, simplifying the message, rendering only its gist, omitting or adding information, or otherwise repurposing the source content. Such outputs, while possibly effective for their intended purpose, diverge from the conventional goal of translation -- namely, to render the source content faithfully using the most appropriate means of the target language. This goal is most straightforward when the source and target texts serve the same communicative purpose and their intended audiences are broadly comparable in the socio-demographic characteristics relevant to the communication. When either the audience or the communicative purpose changes, adaptation of the message itself may become necessary.

At the same time, both the strength and the linguistic manifestation of translationese vary substantially. They are conditioned by multiple factors, most prominently translation direction (i.e. source–target pairing)~\cite{Evert2017, Kunilovskaya2021lit}, register~\cite{Redelinghuys2016, Kunilovskaya2021regs}, translator expertise~\cite{Sutter2017, Bizzoni2021}, and mode (written vs. spoken). Extensive research has also documented recurrent translationese patterns including simplification~\cite{Corpas2008}, shining-through~\cite{teich2003cross}, explicitation~\cite{blumkulka1986shifts, baker1993corpus}, and over-normalisation~\cite{HansenSchirra2011}.

Despite this wealth of descriptive findings, the field still lacks a principled account that integrates these factors within a unified explanatory framework. We therefore ask whether translationese can be understood as a consequence of cognitive constraints inherent in the translation task, and whether an information-theoretic perspective can provide such a framework. 

Building on prior applications of information theory (IT) to language~\cite{Shannon1948, Levy2008}, we adopt the view that linguistic encoding is shaped by cognitive constraints: speakers select forms that balance communicative goals against processing effort~\cite{Gibson1998, jaeger2010redundancy}. We extend this perspective to translation, conceptualising it as a task in which production choices are similarly governed by cognitive pressures. Supporting evidence comes from~\citet{LimEtAl2024}, who show that IT-based difficulty indicators, comparable to those used here, correlate with behavioural measures of translation processing, including production duration and eye-tracking metrics, in an experimental setting.
Motivated by this work, we ask whether such indicators can also account for systematic properties of translated texts. Theoretically, more complex source texts should yield more deviant translations, particularly in simultaneous interpreting. Translationese can thus be seen as a rational trade-off between accuracy and fluency, aimed at managing cognitive load under the assumption that producing translationese-marked text may require less effort. This perspective unifies a range of recurrent deviations (explanatory wordiness, simplification, overly literal renderings, or reduced lexical variety) as cognitively rational strategies rather than lapses in translation quality.

Given the differences between monolingual and cross-lingual language processing, any explanation of translated language must consider the source language input and the cognitive demands of cross-linguistic transfer. Early corpus-based research on translationese, especially following~\citet{baker1993corpus}, focused primarily on target-language output, often without directly incorporating the corresponding source text into the analysis. Subsequent work has increasingly considered the source, although its influence and the cognitive cost of mapping between languages have often been examined indirectly, for example, by comparing translations from multiple source languages into a single target language \citep[e.g.][]{Evert2017, Nikolaev2020, Hu2021}.
In this work, we directly connect the presence of translationese in high-quality professional translations and simultaneous interpreting to the difficulty of the translation task, estimated from aligned source–target segments.\footnote{``Segment'' refers to an aligned unit in the parallel corpus, which may consist of one or more sentences.} We investigate two questions: (1) to what extent translationese can be explained by task difficulty, and (2) how much IT measures of difficulty contribute relative to established structural features. In doing so, we address a gap in previous studies, which rarely analyse translationese in relation to the aligned source material, and test the hypothesis that translationese can be understood as a cognitive response to variation in translation difficulty.

The remainder of the paper is organised as follows. Section~\ref{sec:meth} presents the methodology, including the \textit{translatedness score}, a measure of translationese, and representations of \textit{translation task difficulty}. Section~\ref{sec:exp} describes the experimental setup, covering data, feature handling, and modelling approach. Section~\ref{sec:res} reports the results and discusses their implications, and Section~\ref{sec:fin} offers concluding remarks.

\section{\label{sec:meth}Methodology}
Investigating the link between translation task difficulty and translationese involves two challenges: to quantify translationese and to identify relevant translation task parameters. The next two sections address these in turn.%\footnote{To facilitate reproducibility, the code and experimental data will be released upon acceptance.}
\footnote{To ensure reproducibility, the code and data is released at \nolinkurl{https://github.com/SFB1102/b7-rational-mediation}.}

\subsection{\label{ssec:response}Translatedness Score}  
Previous work demonstrated that humans, including translation experts, struggle with distinguishing translated and original documents~\cite{Baroni2006,wein-2023-human}, let alone quantifying translationese in a document. To address this limitation, we introduce the \textit{translatedness score} -- a measure that captures the extent to which a text exhibits properties of mediated language. This score is defined as the \textit{probability of being a translation}, predicted by a binary classifier trained to distinguish the text types: translations (targets, tgt) vs. comparable originals (org) in the target language.  

A central design requirement is construct validity: the classifier must meaningfully capture translatedness. This entails choosing both an appropriate feature set and an appropriate level of analysis (segment vs. document level).
Translationese indicators should be structural (e.g. morphosyntactic patterns, abstract lexical cues) rather than surface features to avoid exploiting domain- or topic-related signals. Character n-grams~\cite{Popescu2011} or neural embeddings~\cite{Borah2023} were shown to capture the properties of texts that are not directly related to translation-specific linguistic patterns.

Translationese emerges more clearly over larger spans of text. Previous studies employing hand-engineered features (most notably, part-of-speech n-grams) have used chunks ranging from 80 to 2000 words in size~\cite{Volansky2015,rabinovich2015unsupervised}. For instance, for translated English and German parliamentary speeches,~\citet{Kunilovskaya2024} obtained 80.0\% and 88.8\% accuracy on full speeches of approximately 700 tokens using a broad set of delexicalised structural features. The best result achieved in~\citet{Pylypenko2021} on handcrafted features (PoS-trigrams) is 76.6\% and 73.1\% for the same target languages. 

Although segment-level translationese classification is challenging due to a weaker signal, it presents a practical compromise. First, our translation difficulty indicators are mostly defined at the segment level, and aggregating them to larger text spans risks smoothing over differences between segments. Second, assigning a document-level translatedness score to each segment is suboptimal: it increases the number of observations but violates the assumption of independence, as all segments from the same document inherit the same score regardless of their individual properties. Finally, our focus lies in quantifying the extent of translatedness, rather than maximising classification accuracy.
Preliminary experiments confirmed these concerns: both document- and segment-level regression experiments using probability-based scores from document-level classifiers produced noisy results. Based on these considerations, we opted for segment-level classification using the feature set from~\citet{Kunilovskaya2024} with minor modifications. Their feature set covers 58 linguistically-motivated phenomena, including grammatical forms, morphological word classes, clause types, syntactic dependencies, word order patterns, discourse elements, and textual measures. 

Generally, the feature extraction is straightforward and involves counting the occurrences of Universal Dependencies (UD) tags of the same name. For example, the value for \textit{nn} feature is obtained by counting tokens marked with the PoS tag \textit{NOUN}, and finite and past-tense verb forms \textit{fin, pastv} are counted as the number of tokens marked with the morphological features \textit{VerbForm=Fin} and \textit{Tense=Past}. The features \textit{amod, aux, ccomp, nmod, nsubj} and \textit{obj} are counts of the respective dependency relation tags. Translationese predictors that require additional extraction rules or filtering and emerged as useful in our experiments are described in Table~\ref{tab:fancy_feats} (Appendix~\ref{app:class}).  % shared across the four classifications (two target languages by two modes)

Some known translationese predictors are also informative as indicators of translation task complexity. To ensure the validity of the experimental setup, three typical translationese features -- lexical density, number of clauses, and mean dependency distance -- are deliberately reserved as predictors for subsequent regression experiments. Pairwise collinearity between translationese predictors (measured on targets) and structural source difficulty indicators does not exceed Spearman $\rho=0.7$. % branching <-> mean_sent_wc in written_ende
The initial pool of 55 potential translationese indicators was reduced to 37 after filtering out 18 low-variance features (see Section~\ref{ssec:ml}).
For a comprehensive description and a full rationale behind the feature set, we refer the reader to the original study~\cite{Kunilovskaya2024}. 

\subsection{\label{ssec:predictors}Translation Task Difficulty}
\begin{table*}[!t]
\centering
\textsmaller{
\begin{tabular}{@{}p{1.4cm}p{1.6cm}p{11cm}@{}}
\toprule
\textbf{Category} & \textbf{Feature} & \textbf{Description} \\ \midrule

\textbf{IT} & src\_gpt\_AvS 
& Average surprisal of UD tokens in a source segment, derived from dedicated GPT-2 models (English, German). \\ \midrule

\multirow{4}{=}{\parbox[c]{1.5cm}{\centering\textbf{Grammar}}}
& src\_tree\_depth 
& Maximum dependency tree depth (longest path from root to any leaf), capturing hierarchical syntactic complexity. \\
& src\_branching 
& Average branching factor (number of dependents per head), capturing horizontal syntactic complexity. \\
& src\_mdd 
& Mean dependency distance: linear-to-hierarchical integration effort, reflecting comprehension difficulty~\cite{Jing2015}. \\
& src\_n\_clauses 
& Number of finite and non-finite clauses per segment, approximated via clausal dependencies \textit{csubj, advcl, acl, xcomp, ccomp}. \\ \midrule

\multirow{5}{=}{\parbox[c]{1.5cm}{\centering\textbf{Lexis}}}
& src\_lex\_dens 
& Ratio of unique content words (PoS-disambiguated) to total tokens. \\
& src\_wlen 
& Average word length in characters, excluding punctuation. \\
& src\_mwe 
& Number of multiword expressions (\textit{flat}, \textit{fixed}, \textit{compound}), excluding numerals and proper names. \\
& src\_numerals % headed by \textit{NUM}
& Number of numerals, including single-word and multiword expressions. \\
& src\_propn % headed by \textit{PROPN}. 
& Number of proper names, including single-word and multiword expressions.\\
\bottomrule
\end{tabular}%
}
\caption{\label{tab:src_diff}Source-side features used to model source text comprehension difficulty, grouped by feature type.}
\end{table*}

Translation task difficulty is conceived as comprising two components: source text comprehension effort and source-target transfer effort. Each component is represented by features that can be categorised as IT or structural (grammatical and lexical), depending on their nature.

We begin with the IT features, which provide a principled way to quantify the processing cost associated with encoding and transfer effort. The IT indicators operationalise difficulty by capturing the predictability of linguistic units in context: more predictable items are easier to process and may be reduced or omitted~\cite{delogu2017teasing, rutherford2017systematic}, whereas less predictable (more surprising) items incur higher cognitive effort, correlating with 
longer reading times~\cite{jaeger2010redundancy, hao2020probabilistic} and neurophysiological responses~\cite{Huber2024, Michaelov2024}.
These effects are formalised in surprisal theory~\cite{Hale2001, Levy2008}, rooted in Shannon's notion of information~\cite{Shannon1948}.
Within this framework, the central measure of information is \textit{surprisal}, defined as the negative log probability of a word given its context (Equation~\ref{eq:srp}). 

\begin{equation}
	S(w) = -log_2(P(w | context))
	\label{eq:srp}
\end{equation}  

Recent work derives these probabilities from large language models (LLMs), primarily of the GPT-2-small family, whose estimates correlate best with experimental measures of processing difficulty~\cite{oh2023,shain2024large}.

This study uses \textit{average segment surprisal (AvS)}, computed by averaging word-level surprisals across all Stanza-defined words in a segment. 
AvS derived from monolingual GPT-2 models of the source languages (English or German) represents source comprehension difficulty and reflects the predictability of source text words within the source language. AvS derived from translation-direction-specific neural machine translation (NMT) models represents transfer difficulty and captures the uncertainty associated with source-to-target mapping.
AvS was used in translation studies to correlate the information content of source and target segments~\cite{Kunilovskaya2023parallel}, in clinical linguistics to capture both lexical and structural components of sentence information~\cite{rezaii2023MeasuringSentenceInformation} and in text readability research~\cite{lowder2018LexicalPredictabilityNatural}. % they report a significant positive relationship with Flesch-Kincaid-based text difficulty %(r = .72, p < .001)

Complementing NMT-based surprisal, \textit{entropy of translation solutions} captures uncertainty over translation choices and thus reflects transfer complexity. It quantifies the number and distribution of alternative target-language realisations available for a given source item, estimated from large parallel corpora. This measure has been used as an indicator of translation difficulty in several studies~\cite{carl2016critt, teichTranslationInformationTheory2020, kunilovskaya2025predictability}.
Unlike syntagmatic surprisal, which captures predictability within a sequence, translation entropy is paradigmatic: it captures uncertainty among alternative target-language realisations of the same source item.
The entropy of translation solutions is defined in Equation~\ref{eq:ent}. 
\begin{equation}
	H(T \mid S) = - \sum_{t \in T} P(t \mid S) \log_2 P(t \mid S)
	\label{eq:ent}
\end{equation}
Here, \( H(T \mid S) \) denotes the entropy of the set of translations \( T \) for a source item \( S \). The summation iterates over all possible translation variants \( t \in T \), and \( P(t \mid S) \) is the probability of observing translation \( t \) given the source \( S \). Higher entropy indicates a larger set of similarly probable translation options, implying greater cognitive demand for that source item during source–target transfer.

In addition to IT measures, we model translation difficulty using established structural indicators of source-text complexity. Our feature selection draws on two related strands of previous research. First, source-side complexity features used in the \textit{QuEST++} quality estimation framework~\cite{Specia2018} include type--token ratio (TTR), syntactic tree depth and width, number of clauses, and related measures. Second, research on cognitive load in interpreting points to lexical density, formulaicity (frequency of multiword expressions), subordination, and numerical expressions as potential sources of processing difficulty~\cite{Plevoets2020importedload,kajzer2024fluency}. We additionally include mean dependency distance, which has been associated with cognitive load in interpreting~\cite{Liang2017} and has also proved a strong indicator of translationese~\cite{Fan2019}. Together, these measures form a set of ten features intended to capture source-text comprehension difficulty (the \textit{source} subset), presented in Table~\ref{tab:src_diff}.

% Formulaicity was calculated as the sum of the number of trigrams and tetragrams. Trigrams and tetragrams were operationalized as sequences of three or four words, the joint occurrence rate of which exceeded the cut-off of 100 per million words (Plevoets & Defrancq 2018, 8-9)
% Their findings showed that higher lexical density was linked to increased cognitive load, as indicated by more disfluencies (namely, filled pauses), while higher formulaicity correlated with fewer disfluencies. They also investigated the percentage of subordinations in source texts, confirming that syntactic complexity impacts the interpreter’s output. Surprisingly, the frequency of numbers was not revealed as a cognitive load factor. However, in a later, more fine-grained study, numbers emerged were shown tohave the expected effect~\cite{kajzer2024fluency}.

Importantly, the translationese classifiers and regression models rely on non-overlapping feature sets, derived from the target and source texts, respectively. This separation is critical: if the same features were used in both steps, the translatedness score from classifiers (used as the response variable in regression) could inadvertently reflect text difficulty via a shining-through effect, thus compromising the integrity of the experimental design. 

Cross-lingual transfer difficulty in this study was represented by four features (\textit{transfer} subset): average target segment surprisal, mean word alignment score, entropy of translation solutions per segment and teacher-forced pseudo-BLEU. They are described in Table~\ref{tab:trans_diff}. All of transfer features are LLM-based and/or IT-inspired.

\begin{table*}[!t]
\centering
% The idea is that lower token probabilities reflect higher uncertainty during decoding, which may signal greater cross-lingual transfer difficulty. By averaging surprisal values over words in a segment, this measure provides a coarse-grained indicator of how predictable the target segment is from the model’s perspective. Higher average surprisal values thus represent more complex or less canonical translation mappings, a source of translation difficulty and a cognitive load factor--hypothesised to trigger translationese.
\textsmaller{
\begin{tabular}{@{}lp{13cm}@{}}
\toprule
\textbf{Feature} & \textbf{Description} \\ \midrule

mt\_AvS
& Average target segment surprisal: mean surprisal of UD tokens in a target segment derived from NMT models, conditioned on the full source segment and preceding target context. Higher values indicate less predictable source-target mappings and increased transfer difficulty, hypothesised to trigger translationese. \\ \midrule

mean\_align
& Mean word alignment score: computed using contextualised token representations from \textit{bert-base-multilingual-cased}. Subword alignment is based on dot-product attention (layer~8) with bidirectional softmax filtering ($10^{-2}$), following \textit{AWESoME}~\cite{dou2021word}. Segment-level averages over aligned word pairs; higher scores indicate stronger alignment and lower transfer difficulty. \\ \midrule

tot\_entropy
& Entropy of translation solutions per segment: captures uncertainty in translating source lemmas. Entropy is computed on lemmas (not surface tokens) to account for morphological variation. Translation solution sets use word alignments from both training and test splits of \textit{EPIC–EuroParl–UdS}. Missing values for unaligned words and singletons are replaced with the subcorpus median plus two standard deviations. Higher entropy signals less consistent translation patterns and greater cross-lingual transfer difficulty. \\ \midrule

bleu
& Teacher-forced pseudo-BLEU\footnote{Calculating true BLEU score for generated targets takes an unrealistically long time and is not needed for our purposes.}: segment-level BLEU between reference targets and token-wise \textit{argmax} predictions under gold prefixes, computed with \textit{sacrebleu}. Higher scores indicate more predictable target realisations and lower transfer difficulty. \\

\bottomrule
\end{tabular}%
}
\caption{\label{tab:trans_diff}Transfer features used to model cross-lingual transfer difficulty.}

\end{table*}

The basic feature set (comprising source-text difficulty and transfer difficulty indicators) was extended with eight modified variants. The modifications included (i) AvS versions averaged over content words only (\textit{src\_gpt\_AvS\_content, mt\_AvS\_content, mean\_align\_content, tot\_entropy\_content}), (ii) AvS versions averaged over LLM subwords instead of UD-defined surface tokens (\textit{src\_gpt\_AvS\_subw}, \textit{mt\_AvS\_subw}) and (iii) versions computed over syntactic subtrees rather than individual words, namely subtree-based alignment (cosine similarity between aligned trees; \textit{mean\_cosine}) and entropy over trees (\textit{tot\_entropy\_trees}). Subtrees were extracted using UD tags (id and head\_id), limiting the depths of possible subtrees to 1. 
Taken together, the 22 indicators capture complementary dimensions of translation task difficulty that may shape translation decisions. The alphabetical list appears in Table~\ref{tab:svr_uni} (Appendix~\ref{app:reg}).

The general expectation is that greater difficulty increases cognitive load, under which translators prioritise content preservation over target-language naturalness, resulting in higher translatedness of the output. Accordingly, we expect a positive association with translatedness for features signalling structural and lexical complexity of the source (\textit{src\_gpt\_AvS, src\_branching, src\_lex\_dens, src\_mdd, src\_wlen, src\_n\_clauses, src\_tree\_depth}), translation ambiguity (\textit{tot\_entropy}) and cross-lingual predictability as reflected in MT surprisal (\textit{mt\_AvS}). Features associated with processing ease or the availability of stereotypical, structurally isomorphic solutions -- such as formulaicity (\textit{src\_mwe}), alignment strength (\textit{mean\_align}), and higher MT performance (\textit{bleu}) -- are expected to correlate negatively with translatedness. Two additional indicators, numerals (\textit{src\_numerals}) and proper names (\textit{src\_propn}), are included as potentially relevant predictors, although their effect direction is less clear.

% The experiments used LLM-based features from base and fine-tuned models. Fine-tuning was performed on 120k non-overlapping segments from the \textit{EPIC–EuroParl–UdS} corpus. According to the corpus curators~\cite{kunilovskaya2026perspectives}, fine-tuning was thorough and robust, ensuring reliable surprisal estimates.
% Despite this, downstream regression gains from the fine-tuned models were limited and inconsistent across settings (see Figure~\ref{fig:base_ft}, Appendix~\ref{app:reg}); therefore, all results reported in this study use the base models. 

The experiments used LLM-based features from base and fine-tuned models. Fine-tuning was performed on 120k non-overlapping segments from the \textit{EPIC–EuroParl–UdS} corpus. According to the corpus curators~\cite{kunilovskaya2026perspectives}, fine-tuning was thorough and robust, ensuring reliable surprisal estimates.
Despite this, downstream regression gains from the fine-tuned models were limited and inconsistent across settings (see Figure~\ref{fig:base_ft}, Appendix~\ref{app:reg}); therefore, all results reported in this study use the base models.

\section{\label{sec:exp}Experimental Setup}
\subsection{\label{ssec:data}Data}
The data come from \textit{EPIC-EuroParl-UdS}\footnote{The data used in this study are publicly available as provided by the original authors:
\nolinkurl{https://zenodo.org/records/19010272}.}, a bidirectional English--German corpus containing written translations and simultaneous interpretations of European Parliament debates~\cite{kunilovskaya2026perspectives}. Both components originate from official European Parliament records. The written data are based on the officially published proceedings and their translations, while the spoken data consist of manual transcriptions of official video recordings of parliamentary debates and their simultaneous interpretations.

The corpus is aligned at document and segment levels and balanced across translation directions and modes. It contains UD annotation and word-level surprisal indices from GPT-2\footnote{English~\cite{GPT2-English}, German~\cite{GPT2-German}}, and OPUS NMT de-en and en-de models~\cite{tiedemann-thottingal-2020-opus}. %, pre-trained and fine-tuned on the written originals to adapt the base models to topical domain.

Table~\ref{tab:corpora} summarises the quantitative parameters of the eight text categories by mode (written vs. spoken), language pair (deen, ende), and text type (original/source vs. target). % Note that for translationese classification, segment alignment is not required: source segments from one translation direction serve as comparable originals for target segments in the reverse direction.

\begin{table}[!ht]
\centering
\textsmaller{
	\resizebox{.9\columnwidth}{!}{%
	\begin{tabular}{@{}p{1cm}llrrr@{}}
		\toprule
        mode                & lpair              & type  & words    & segs &  docs\\ \midrule
        \multirow{4}{*}{spoken} & \multirow{2}{*}{deen} & src, org & 62,900  & 3,060& \multirow{2}{*}{165}\\
		&                        & tgt & 60,408 & 2,896 & \\
		& 						\multirow{2}{*}{ende} & src, org & 69,801 & 3,260 & \multirow{2}{*}{137}\\
		&                        & tgt & 60,590  & 3,092 & \\ \midrule

		\multirow{4}{=}{\parbox{1cm}{written}} & \multirow{2}{*}{deen} & src, org & 77,225  & 3,211 & \multirow{2}{*}{170}\\
		&                        & tgt & 85,380 & 3,216& \\
		& \multirow{2}{*}{ende} & src, org & 76,061  & 3,063 & \multirow{2}{*}{170}\\
		&                        & tgt & 76,588  & 3,063 & \\ 
		\bottomrule
	\end{tabular}
	 }}
	\caption{\label{tab:corpora}Quantitative parameters of the textual data by mode, language pair (lpair) and text type.}
	
\end{table}
Average segment length ranges from 25.5 to 27.8 words for written data and 19.5 to 21.4 words for spoken data, with standard deviations of approximately 15 and 13 words, respectively. Segments shorter than 4 tokens were filtered out.

\subsection{\label{ssec:prepro}Feature Preprocessing}
Classification and regression features went through the same data preprocessing pipeline.
Following~\cite{Evert2017}, morphological and lexical raw counts and per-word/subtree alignment scores are normalised by segment word count; syntactic and continuous features (\texttt{mdd}, \texttt{mhd}, \texttt{lex\_dens}, \texttt{ttr}) are averaged across sentences.
To enhance comparability and reduce the influence of outliers, highly right-skewed features were transformed using $log(1 + x)$ function, which compresses long tails and makes distributions more symmetric. Skewness was measured using Fisher's moment coefficient (adjusted for sample size), with a threshold of 1.0 used to identify features requiring transformation.  % Highly skewed feats are log-transformed (9): ['bleu', 'mt_AvS_content']
% short sentences: Vielen Dank! Ganz wichtig! And today? Unfortunately not.

\subsection{\label{ssec:ml}Machine Learning Models}
Translationese classifiers and subsequent regressions were run in the same experimental setup. 
The machine learning (ML) models were trained separately for each language (de = German, en = English) and mode under a 10-fold cross-validation setup. We used a linear \textit{Support Vector Machine} (SVM) with default \textit{scikit-learn} parameters (kernel=`linear', C=1.0). To prevent data leakage, segment-level experiments employed \textit{GroupKFold}, ensuring that all segments from the same document were assigned to the same split, either training or test. 
Classifiers' performance was evaluated using macro-F1 (precision and recall are well-balanced in all experiments). For regression, we report Spearman correlation coefficient ($\rho$) between predicted scores and true translatedness score, the measure of regression fit ($R^2$) and Mean Absolute Error (MAE).

Many of the count-based features are susceptible to \textit{low variance}, particularly at the segment level, making them noisy and uninformative for ML algorithms. Consequently, we filtered out the 18 lowest-variance features shared across all mode-language datasets. 
This global decision reduced models' complexity and fixed the input dimensionality at 37 features across classifiers, facilitating direct comparison of their performance. None of the difficulty features had low variance.

Additionally, separately for each mode and target language, we eliminated \textit{collinear features} (Pearson’s $r$ > 0.85), retaining the feature with the higher univariate F1 score to preserve non-redundant signal.  % non-arbitrary removal! 
All features were independently scaled to zero mean and unit standard deviation. 

Feature selection was performed using cross-validated \textit{Recursive Feature Elimination} (RFECV) from \textit{scikit-learn} (folds=10), which selects features iteratively within each fold. A minimum of two features was enforced. The reported performance after feature selection should be interpreted as conservative, as RFECV favours stable individual predictors and does not capture potential joint effects among weaker features. %As a greedy approximation to an ablation study, the algorithm removes the least informative features at each step based on current weights, without exhaustively evaluating all possible feature combinations.

To interpret feature contributions in the final classification and regression models, we employed \textit{SHapley Additive exPlanations} (SHAP)~\cite{lundberg2017unified}, a unified framework based on Shapley values from cooperative game theory, which quantifies the average marginal contribution of each feature to the model's predictions across all possible feature coalitions.  % SHAP calculates the feature's contribution when it's added to a group that already contains "Feature A." "marginal" refers to the change in an outcome when a specific feature joins a group, for all possible groups.

\section{\label{sec:res}Results}

\subsection{\label{ssec:class}Translationese Classification}
Table~\ref{tab:clf} reports macro-F1 scores averaged across folds for the four segment-level classification tasks (with standard deviations), alongside document-level results for comparison. 

\begin{table}[!h]
\centering
\textsmaller{
	\begin{tabular}{@{}lrcrr@{}}
		\toprule
		mode & support & lang & F1$\pm$SD& selected \\ 
		\midrule
		\multicolumn{5}{l}{segment-level} \\ 
		\midrule
		\multirow{2}{*}{written}& 6,274  & de   &   60.85$\pm$2.61& 13 \\
		                         & 6,279  & en    & 55.74$\pm$2.94& 33  \\ 
		\multirow{2}{*}{spoken}& 6,152  & de    & 60.44$\pm$1.62& 34 \\
		                     & 6,156  & en    & 59.50$\pm$3.49& 12  \\ 
		\midrule
		\multicolumn{5}{l}{document-level} \\
		\midrule
		\multirow{2}{*}{written}& \multirow{2}{*}{340}  & de     & 82.27$\pm$6.34& 21     \\
		                     &   & en     & 74.29$\pm$5.89& 18     \\ 
		\multirow{2}{*}{spoken}& \multirow{2}{*}{302}  & de    & 84.02$\pm$5.91& 27      \\
		                     &   & en     & 76.30$\pm$6.91& 26  \\
		\bottomrule
	\end{tabular}%
    }
	\caption{\label{tab:clf}Performance of segment- and document-level SVM classifiers. \textit{Selected} reflects the final feature subset after collinearity removal and RFECV (out of 37).}%
    
\end{table}

% The performance of the segment-level classifiers is relatively low, with F1 scores around 60\%, only 10 percentage points above the random baseline for a balanced binary task. The lower performance compared to document-level classification (F1 scores between 74\% and 84\%) is expected, given the weaker translationese signal in shorter text spans. The signal is further reduced by excluding lexical density and two syntactic complexity indicators from the translationese feature set.

% Nevertheless, we argue that the classifier's probabilities for the translated class meaningfully reflect the degree of translatedness: even if a translation is predicted as an original, the lower translatedness score is a useful signal. In fact, a highly confident classifier would produce a left-skewed probability distribution, limiting variability and making the scores less informative as a gradient measure of translatedness.
% In our experiments, the probabilities are normally distributed, with a subtle shift to the right of the 0.5 threshold, reflecting more salient or uniform patterns in mediated language (see Figure~\ref{fig:probs}, Appendix~\ref{app:reg}). 

The performance of the segment-level classifiers is relatively low, with F1 scores around 60\%, only 10 percentage points above the random baseline for a balanced binary task. The lower performance compared to document-level classification (F1 scores between 74\% and 84\%) is expected, given the weaker translationese signal in shorter text spans. The signal is further reduced by excluding lexical density and two syntactic complexity indicators from the translationese feature set.

Nevertheless, our goal is not to maximise classification accuracy, but to capture graded variation in translatedness using theoretically motivated structural features. We argue that the classifier's probability for the translated class provides such a gradient: even when a translation is classified as original, a lower translatedness score can meaningfully indicate a weaker translationese signal. For comparison, we additionally tested embedding-based classifiers (DeBERTa and DistilBERT variants), which achieve over 96\% document-level accuracy. While expected, such high performance is not necessarily advantageous for our purposes: embedding-based models may exploit signals not directly related to translationese~\citep{Borah2023}, and highly confident predictions reduce the variation in translatedness scores that constitutes our response variable. Our structural classifiers instead capture a weaker but interpretable signal under conditions explicitly designed to avoid information leakage. In our experiments, the resulting probabilities are approximately normally distributed, with a subtle shift to the right of the 0.5 threshold, reflecting more salient or uniform patterns in mediated language (Figure~\ref{fig:probs}, Appendix~\ref{app:reg}). These patterns are consistent with previous findings of higher predictability of simultaneous interpreting transcripts and greater detectability of mediated German compared to English~\citep{Evert2017}.

We analyse deviation patterns using feature importances. Feature-level patterns are more reliable at the document level, as segment-level signals are weakened and distorted; this is visible in Figure~\ref{fig:shared_svm}, where SHAP profiles reveal different feature strength and groupings across representations. The results further suggest that translationese is captured primarily by feature combinations rather than strongly deviant individual cues. Only a few features show notable univariate performance, especially in written data (Table~\ref{tab:doc_svm_arrows}, Appendix~\ref{app:class}). As shown in Figure~\ref{fig:doc_svm_shap_beeswarm}, features without statistically significant class differences can still contribute to predictions; for example, lower frequencies of personal pronouns (\textit{ppron}) favour the target class in spoken English (subplot~d), despite the non-significant difference in Table~\ref{tab:doc_svm_arrows}.

\subsection{\label{ssec:reg}Regression Results and Discussion}
This section tests whether translation task difficulty predicts translatedness score from the classifiers and compares the explanatory power of two difficulty components (source comprehension vs.\ cross-lingual transfer) and two types of indicators (IT vs.\ structural).

\begin{figure}[!h]
    \centering
    % First Subfigure
    \begin{subfigure}{.9\columnwidth}
        \centering
        \includegraphics[trim=0 10 0 0, clip, width=\linewidth]{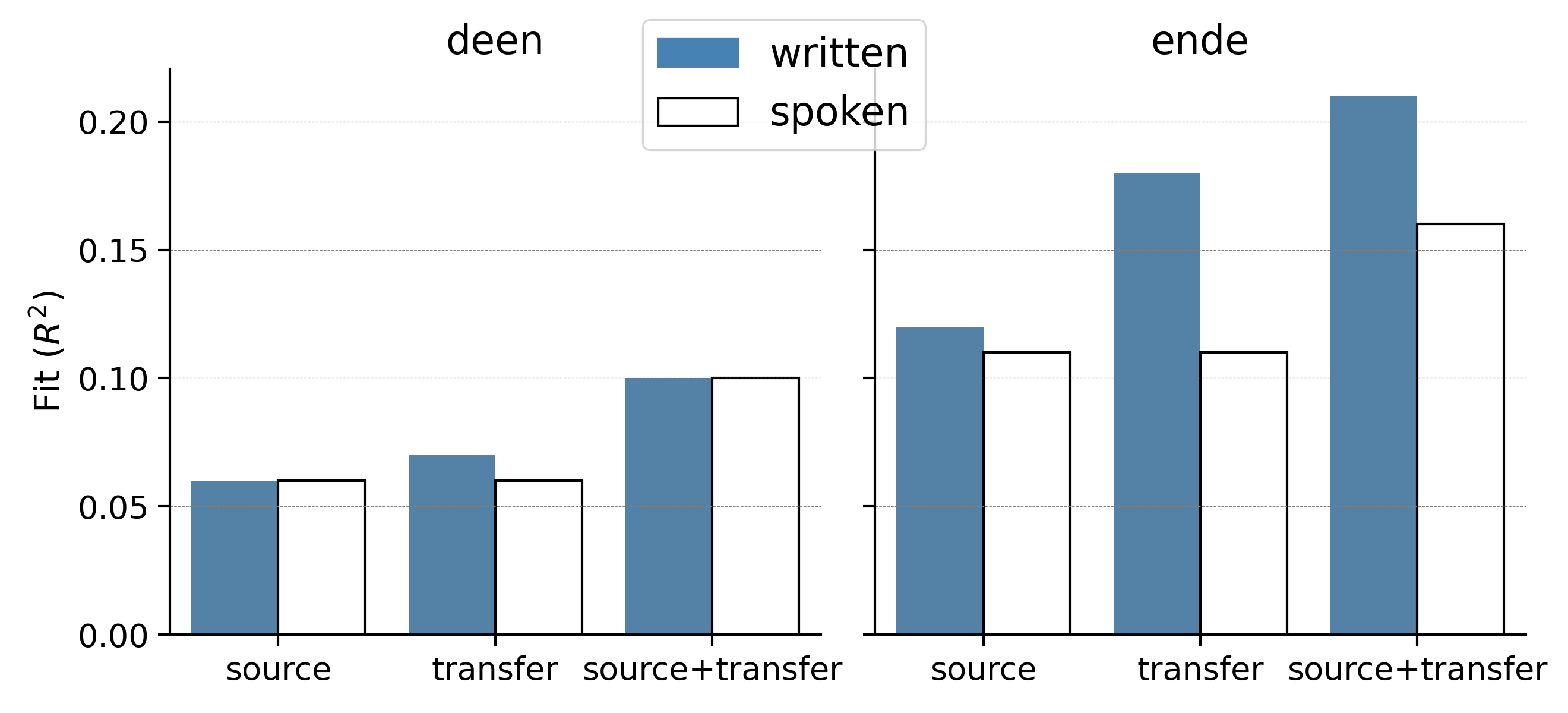}
        \caption{\label{fig:trends-transfer}Source-based and transfer-specific predictors.}
    \end{subfigure}

    \vspace{.5em} % Adds vertical spacing between the two images

    % Second Subfigure
    \begin{subfigure}{.9\columnwidth}
        \centering
        \includegraphics[trim=0 10 0 0, clip, width=\linewidth]{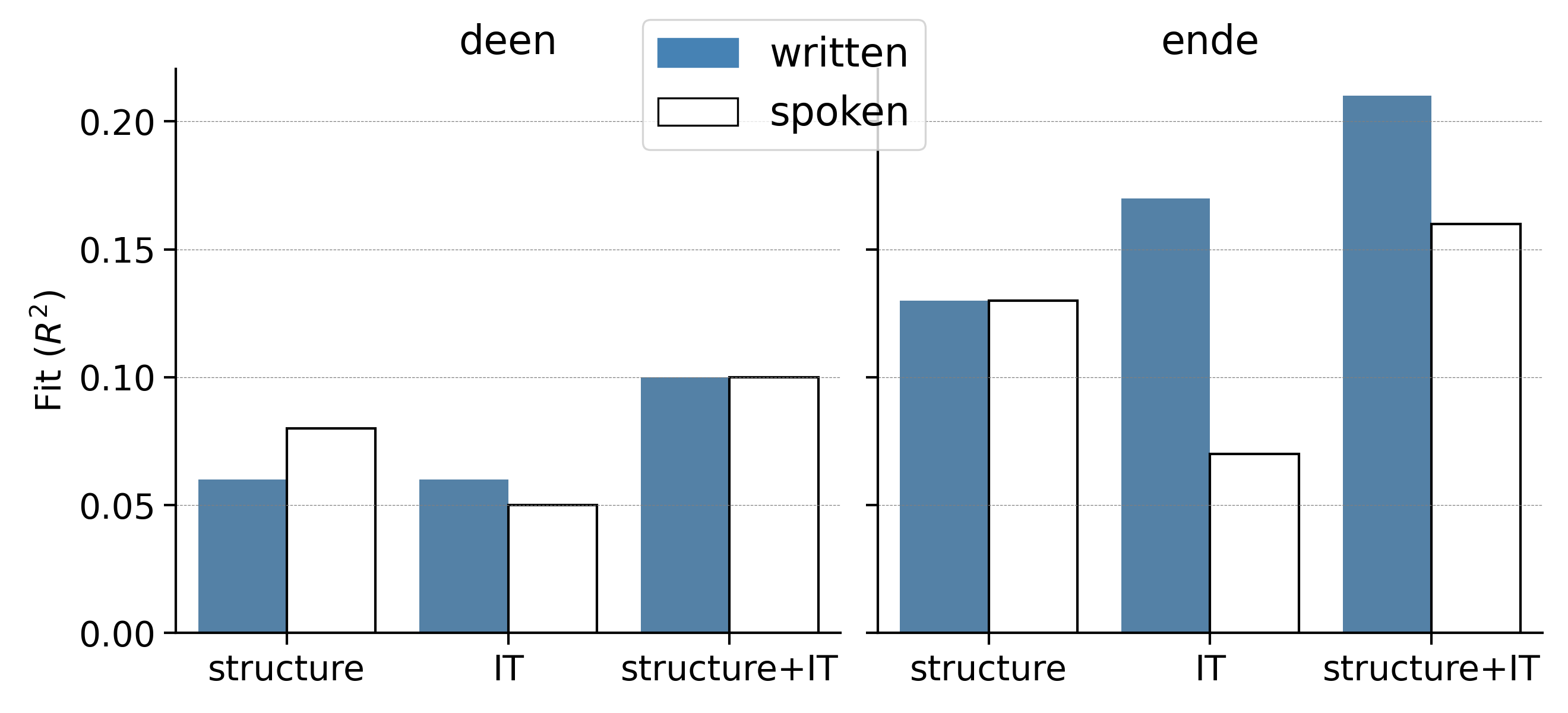}
        \caption{\label{fig:trends-structural}Structural and information-theoretic predictors.}
    \end{subfigure}

    \caption{\label{fig:trends}Variance in translatedness explained by linear regression models across language pairs and mediation modes. The two panels compare difficulty components and indicator types on the x-axis.}
\end{figure}

For ease of interpretation, $R^2$ values are visualised in Figure~\ref{fig:trends}, while the full tabular results are provided in Appendix~\ref{app:reg} (Tables~\ref{tab:svr_diff} and ~\ref{tab:svr_feat_type}).
The regression results show that translation task difficulty explains a modest but consistent share of the variance in translatedness scores, with clear differences by direction and mode.
Overall, the models account for between 6\% and 21\% of the variance. 
The explanatory power is highest for English$\rightarrow$German direction, particularly in the written mode, where the combined model reaches $R^2 = 0.21$. In contrast, the models explain considerably less variance for German$\rightarrow$English, where the best-performing models in both written and spoken modes reach only $R^2 = 0.10$, with smaller differences between the modes.

Across all settings, the combined \textit{source+transfer} models consistently outperform models based on a single difficulty type, indicating that comprehension-related and transfer-related factors capture complementary aspects of translation task difficulty. Transfer features tend to be more predictive of translationese than source complexity in written mode, but this edge disappears in interpreting, where the difficulty components are on par in both translation directions (Figure~\ref{fig:trends}a). 

With regard to the type of difficulty indicators (Figure~\ref{fig:trends}b, Table~\ref{tab:svr_feat_type}), the two feature types (IT vs. structural) perform comparably in German$\rightarrow$English written translation ($R^2=0.06$). For English$\rightarrow$German translation, IT features clearly outperform structural alternatives ($R^2=0.17$ vs.\ $0.13$). In spoken mode, however, structural features yield substantially stronger results in both translation directions.

\begin{figure}[!h]
	\centering  
	\includegraphics[width=\columnwidth]{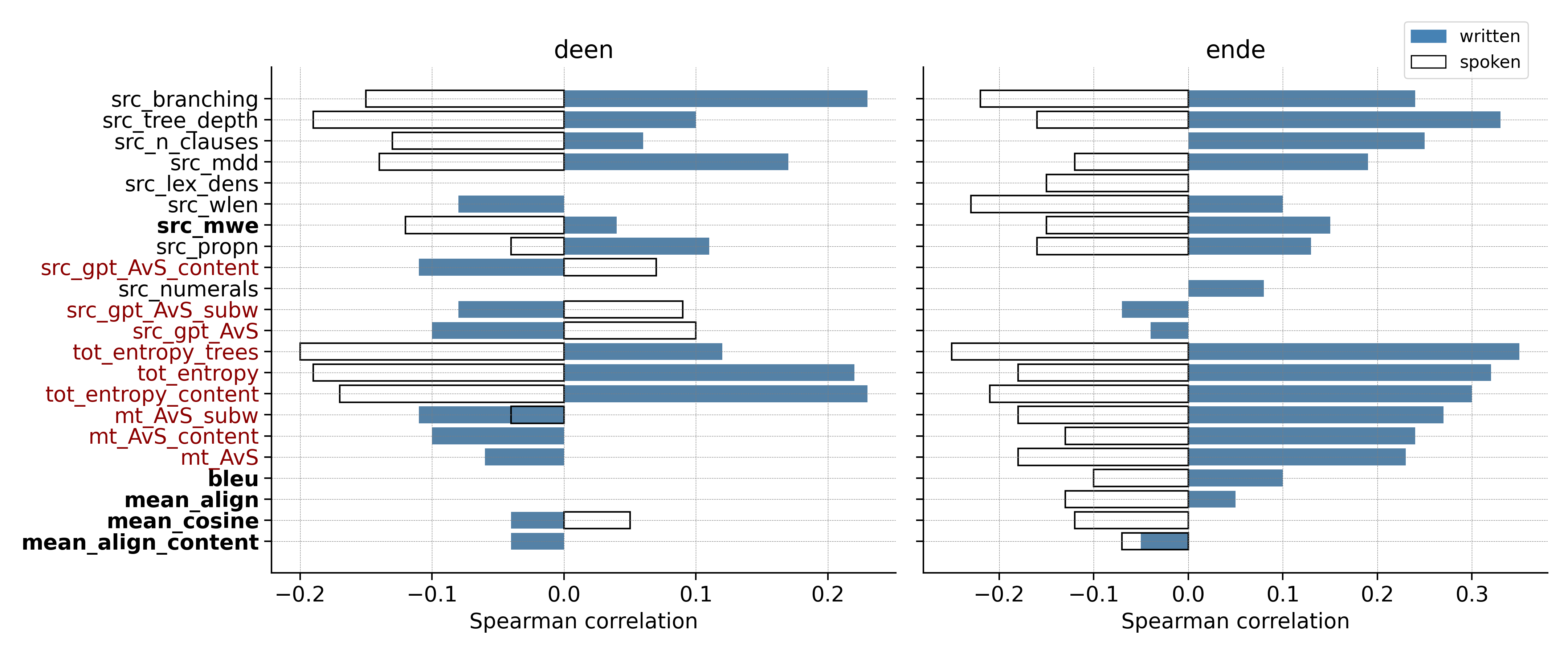}
	\caption{\label{fig:corrs}Univariate analysis: Spearman correlation between translation difficulty indicators and translatedness score. 
    Source text difficulty indicators (src\_) appear first. Features expected to correlate negatively with translatedness are shown in bold, and IT features are in red on the y-axis. The bars for non-significant $\rho$ values are removed.}
\end{figure}

An unexpected finding emerged in the analysis of the directionality of individual feature contributions. 
Figure~\ref{fig:corrs} shows Spearman correlations between translation-difficulty features and translatedness for written (blue bars) and spoken (empty bars) data in both translation directions, revealing marked cross-modal asymmetries. For several features, written and spoken data show opposite tendencies: notably, \textit{src\_branching} and \textit{src\_tree\_depth} correlate positively with translatedness in written tasks but negatively in spoken tasks across both language pairs. 
One possible explanation is that written translation allows greater preservation of source structure, so increasing source complexity may produce stronger source-language interference and hence more translationese. Under the time constraints of simultaneous interpreting, however, greater source complexity makes close structural reproduction increasingly difficult. Incremental processing may instead promote simplification and reformulation in target-language-preferred forms, resulting, paradoxically, in less translationese as source complexity increases.

This interpretation, however, requires some qualification with respect to translation direction. 
In German-to-English interpreting, structural source features remain more influential than information-theoretic features (see the taller transparent ``spoken'' bar in Figure~\ref{fig:trends-structural}), suggesting that the structural properties of spoken German may still impose substantial processing demands. 
Indeed, although spoken source texts are structurally simpler than written ones in both languages, the difference is systematically smaller for German; German source texts also show greater complexity than English on mean dependency distance and word length (\texttt{src\_mdd} and \texttt{src\_wlen} in Table~\ref{tab:svr_uni}, compare \texttt{written\_deen|mean} with \texttt{written\_ende|mean} and \texttt{spoken\_deen|mean} with \texttt{spoke\_ende|mean}). Source structure may therefore remain more consequential when interpreting from German, whose spoken input retains more written-like structural complexity. Whether this reflects the style of German parliamentary speech or broader cross-linguistic structural differences remains an open question.
% This pattern is compatible with different responses to source complexity across modes: written translation allows greater preservation of source structure, whereas simultaneous interpreting prioritises rapid transfer of content under time constraints and may therefore involve greater structural reformulation. Thus, increasing source complexity need not result in greater structural interference in interpreted output.

% This suggests that while written translationese tends to mirror source complexity, spoken mode may elicit the opposite response: the more difficult the task, the more fluent and less translationese-marked the output. 
These findings are further supported by the SHAP importance rankings and granular statistical analysis provided in Figures~\ref{fig:shared_seg_svr} and~\ref{fig:svr_shap_beeswarm} Appendix~\ref{app:reg}.

The heatmap in Figure~\ref{fig:shared_seg_svr} shows that IT indicators of transfer difficulty, notably \textit{mt\_AvS} and \textit{translation solution entropy}, are the strongest predictors of translationese, particularly in the English$\rightarrow$German direction. In contrast, the lighter colour intensity for the German$\rightarrow$English direction indicates a weaker association between translation task difficulty and translationese, consistent with the generally lower intensity of translationese in this direction. 

\section{\label{sec:fin}Conclusion}

This study examined translationese as a linguistic marker of cognitive processes involved in human translation and interpreting. By linking measures of translation task difficulty to the degree of translatedness in the target text, we explored whether systematic deviations in mediated language can be explained by the complexity of the translation task.

The results show that translation difficulty accounts for a measurable but limited share of variation in translationese. Regression analyses explain up to 21\% of the variance in translatedness scores, with the relationship strongest in written translation and particularly in the English$\rightarrow$German direction. Cross-lingual transfer indicators generally provide stronger explanatory signals than source-comprehension measures, suggesting that transfer difficulty contributes to translationese but cannot provide a comprehensive account of it.

The relationship also differs across mediation modes. Increasing source complexity tends to correlate with more translationese in written output but, for several structural features, with less translationese in interpreting, possibly reflecting greater simplification and target-oriented reformulation under time constraints. This pattern is direction-dependent, with source structure remaining comparatively influential in German$\rightarrow$English interpreting. These exploratory findings point to different responses to translation difficulty across modes and languages.

Importantly, information-theoretic measures do not consistently outperform traditional structural indicators and have particularly limited explanatory power for spoken data. Thus, under the present operationalisation, surprisal- and entropy-based metrics appear too coarse to support a unified information-theoretic account of translationese. Instead, our results delimit its empirical scope: cognitive difficulty contributes to translationese, but its explanatory power depends on modality, translation direction, and its operationalisation.

More broadly, by relating gradient translatedness to difficulty estimated from aligned source--target data, this study provides a direct empirical test of cognitive explanations of translationese and quantifies their explanatory limits. The results motivate more targeted models that distinguish sources of difficulty and the strategies translators and interpreters use to manage them.

\section*{Limitations}
Several limitations should be considered when interpreting the results.
First, the outcomes depend on the chosen feature set. Linguistic feature extraction is inherently fragile and sensitive to preprocessing decisions, annotation errors, and normalisation choices. Many frequency-based features are also sensitive to sample length, which may affect their stability across segment and document representations.

Second, latent structural dependencies between source and target features may introduce a degree of indirect information leakage, potentially confounding the observed effects of source comprehension on translationese.

Third, the analysis does not explicitly model individual speaker effects. Especially in spoken data, stylistic variation between speakers may influence feature distributions and interact with mediation-related patterns. Future work should therefore account for speaker identity and other individual-level variables.

Fourth, the results are derived from a specific corpus and set of language pairs. While the observed patterns are consistent within this dataset, their generalisability to other domains, genres, or language combinations remains to be tested. Although the analysis suggests systematic differences between written and spoken mediation, these patterns may also reflect differences in topics, communicative conditions and document length across spoken and written subcorpora, although they come from the same global source.

Finally, this study does not attempt a detailed linguistic analysis of the individual predictors and their interactions. Although the modelling results reveal systematic patterns in the feature space, a closer examination of how specific linguistic cues contribute to these patterns would provide additional explanatory depth. Conducting such analysis is left for future work.

We do not foresee direct negative societal impacts from this study. Nevertheless, metrics of translation difficulty or predictability could be misapplied in evaluative settings (e.g., assessing translators or systems) if treated as definitive indicators rather than exploratory analytical tools.

\section*{Acknowledgments}
We are deeply grateful to Christina Pollkläsener for her enduring support and many insightful discussions throughout this project, as well as for her careful review and proofreading of the final manuscript. We also thank the anonymous reviewers for their thoughtful and constructive feedback, which helped us substantially improve the paper.

This work was funded by the Deutsche Forschungsgemeinschaft (DFG, German Research Foundation), Project-ID 232722074, as part of the Collaborative Research Center SFB 1102.

\bibliography{subs}

\appendix
\clearpage
\onecolumn
\raggedbottom
\section{\label{app:class}Appendix. Additional Details on Segment-Level Translationese}
This appendix provides details on features and their performance, beginning with a description of some complex predictors. We then look into feature impact through SHAP-based importance rankings and class-prediction contributions, and directional frequency comparisons, complemented with univariate discriminative performance and statistical analysis of differences.

\vspace{2em}
\noindent
\begin{minipage}{\linewidth}
\centering

\textsmaller{
\begin{tabular}{@{}p{1.6cm}p{13cm}@{}}
\toprule
\textbf{Feature} & \textbf{Description} \\ \midrule
advmod & Adverbial modifiers per sentence, excluding negatives. Count of \textit{advmod} dependency relations for tokens where morphological features do not contain \textit{=Neg}. \\
mean\_sent\_wc & Mean sentence length. Count of UD tokens per sentence. \\
mhd & Mean hierarchical distance. Average length of paths from the root to all nodes along the dependency edges~\cite{Jing2015}. \\
nnargs & Ratio of nouns, inc. proper names, as core verb arguments to the total of arguments. Count of tokens tagged as \textit{NOUN} or \textit{PROPN} to tokens tagged \textit{nsubj}, \textit{obj}, or \textit{iobj}.\\
relcl & Relative clauses. Count of relative clauses in declarative sentences based on language-specific cues. For English, count of pronouns such as \textit{which, that, whose, whom, what, who} tagged as pronouns with head relation \textit{acl:relcl}. For German, count of pronouns like \textit{der, welch, was, wer}, or pronouns containing \textit{wo} with feature \textit{PronType=Int,Rel} occurring near commas (up to 3 tokens away). \\
epist & Discourse markers of epistemic stance. Count of total occurrences of items from pre-defined language-specific lists (64 items for English, 74 for German, e.g., \textit{at least, perhaps, angeblich, kein Zweifel}). \\
mpred & Modal predicates, including modal adjectives in predicative function (e.g., \textit{likely, obvious, supposed, klar, unbedingt, möglich}). For English, count of  modal auxiliaries (PoS tag \textit{MD}) excluding \textit{will} and \textit{shall}, including non-\textit{AUX} \textit{have} followed by nearby infinitive verbs, ignoring causatives, plus modal adjectives with \textit{AUX} dependents. For German, count of core modal verbs (\textit{dürfen, können, mögen, müssen, sollen, wollen}) and modal adjectives in predicative function. \\
ppron & Personal pronouns. Count of personal pronouns in a segment, filtered by language-specific pronoun lists and morphological features: \textit{PRON} with \textit{Person=} feature, excluding possessive pronouns (\textit{Poss=Yes}). \\
\bottomrule
\end{tabular}%
}
\captionof{table}{\label{tab:fancy_feats}Examples of useful translationese predictors requiring less-trivial extraction rules and/or filtering.}

\end{minipage}
\vspace{2em}
\noindent
\begin{figure}[!h]
    \centering
    % Left Subfigure
    \begin{subfigure}{0.49\linewidth}
        \centering
        \includegraphics[trim=0 10 0 5, clip, width=\linewidth]{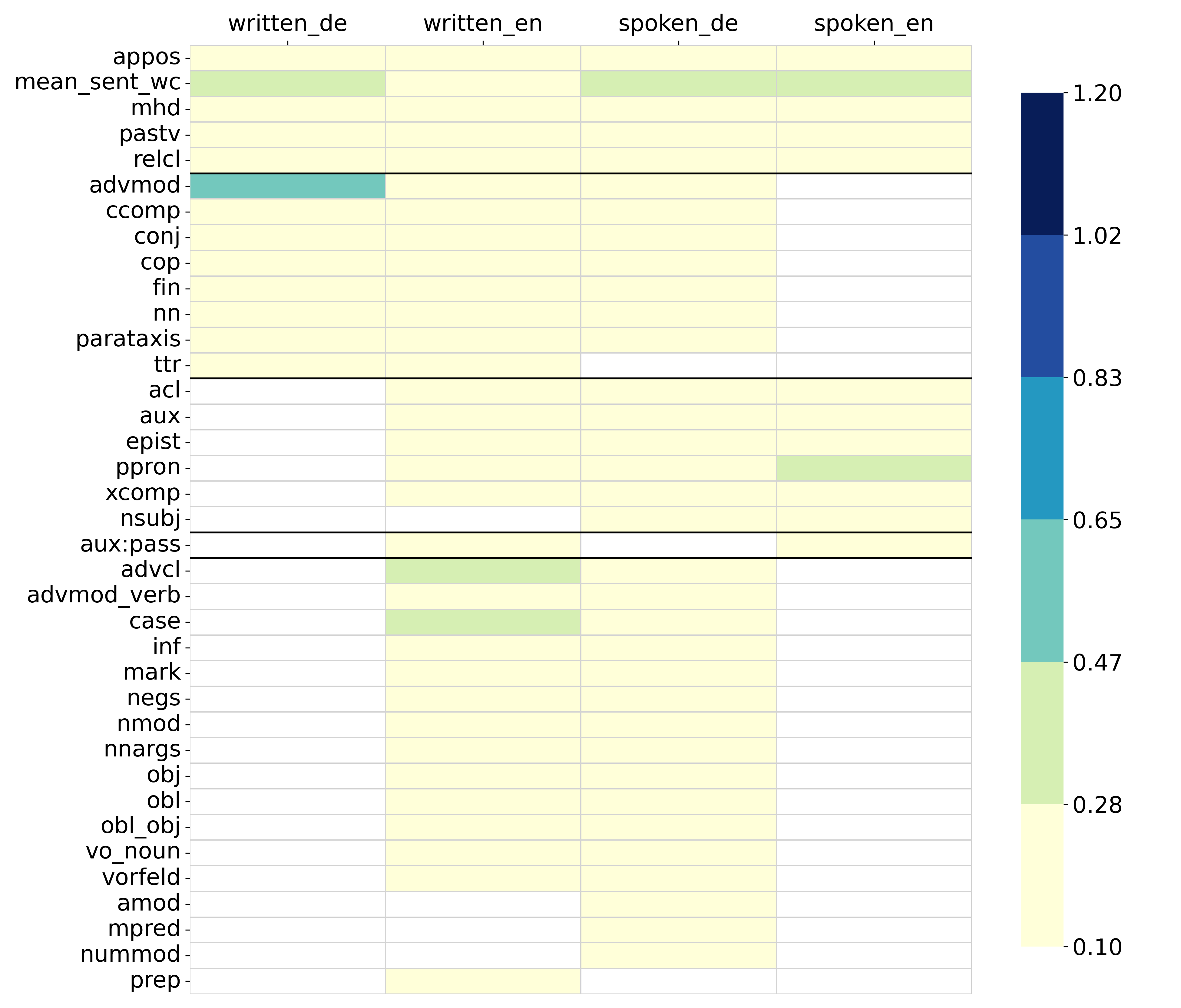}
        \caption{Segment-level features.}
        \label{fig:shared_seg_svm}
    \end{subfigure}
    \hfill % Adds spacing to push subfigures to the edges
    % Right Subfigure
    \begin{subfigure}{0.49\linewidth}
        \centering
        \includegraphics[trim=0 10 0 5, clip, width=\linewidth]{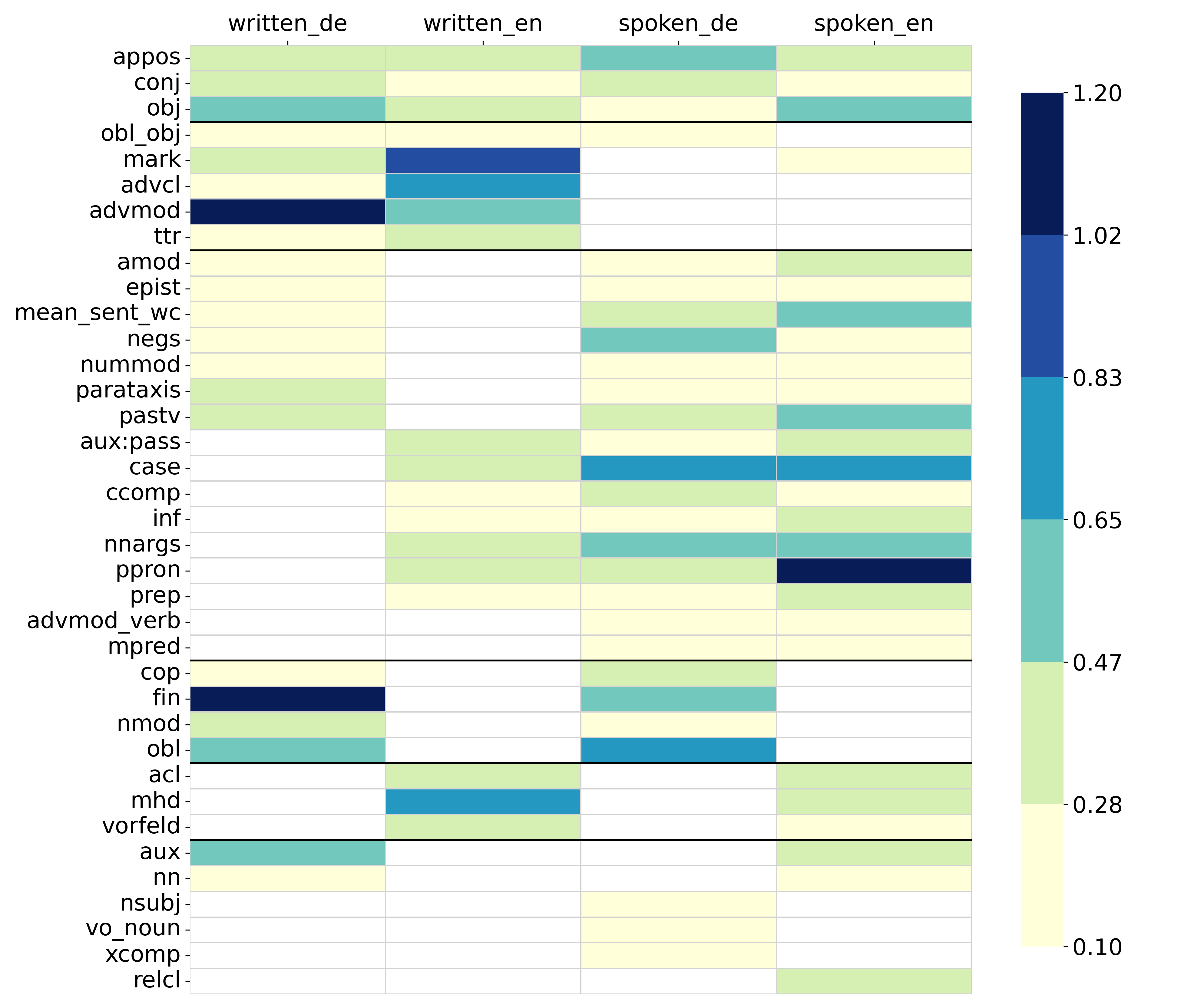}
        \caption{Document-level features.}
        \label{fig:shared_doc_svm}
    \end{subfigure}

    \vspace{10pt} % Space between images and main caption
    
    \caption{\label{fig:shared_svm}Segment vs. document feature importances: Average absolute SHAP values for translationese predictors by mode and target language on a shared scale. Darker shades indicate higher feature importance; blank cells indicate that a feature was not selected for the given task. Features are grouped by selection overlap: the top block shows predictors selected by all four classifiers, followed by predictors shared by mode (written, spoken) and by target language (German, English). Segment-level analysis dilutes feature importance and obscures patterns that are clearer at the document level. The grouping sizes further suggest that mode yields more consistent deviation patterns than translation direction.}
\end{figure}

\noindent
\begin{minipage}{\linewidth}
\centering
	\begin{tabular}{@{}ll|cccc|cccc@{}}
\toprule
		\multirow{2}{*}{\#} & \multirow{2}{*}{feature}       & \multicolumn{2}{c}{written\_de}  & \multicolumn{2}{c|}{written\_en}  & \multicolumn{2}{c}{spoken\_de}  & \multicolumn{2}{c}{spoken\_en}  \\ 

        \cmidrule(lr){3-4}
\cmidrule(lr){5-6}
\cmidrule(lr){7-8}
\cmidrule(lr){9-10}
        
		&        & tgt$\rightarrow$org & F1\_svm & tgt$\rightarrow$org & F1\_svm & tgt$\rightarrow$org & F1\_svm & tgt$\rightarrow$org & F1\_svm \\ \midrule
1 & acl & $\nearrow$ & 58.9 & -- & 45.1 & -- & 51.1 & $\searrow$ & 62.2 \\
2 & advcl & $\nearrow$ & 55.1 & -- & 49.8 & -- & 46.5 & $\searrow$ & 61.2 \\
3 & advmod & $\searrow$ & \textbf{65.5} & $\nearrow$ & 59.1 & -- & -- & $\searrow$ & 56.4 \\
4 & advmod\_verb & -- & 52.1 & -- & 39.6 & -- & 44.1 & -- & 56.1 \\
5 & amod & $\nearrow$ & 56.0 & -- & 48.7 & $\searrow$ &  \textbf{71.3} & $\searrow$ & 53.5 \\
6 & appos & $\searrow$ & 52.7 & $\nearrow$ & 47.9 & $\searrow$ & \textbf{66.5} & -- & 42.3 \\
7 & aux & -- & 40.6 & $\searrow$ & 53.1 & $\searrow$ & 57.0 & -- & -- \\
8 & aux:pass & $\nearrow$ & 58.2 & $\searrow$ & 54.2 & -- & 45.5 & -- & 38.6 \\
9 & case & $\nearrow$ & 63.7 & -- & 46.5 & $\searrow$ &  \textbf{68.3} & $\searrow$ &  \textbf{66.0} \\
10 & ccomp & $\nearrow$ & 55.1 & -- & 43.2 & -- & 44.1 & $\searrow$ & 58.0 \\
11 & conj & -- & 50.6 & $\nearrow$ & 55.7 & $\searrow$ &  \textbf{70.5} & $\searrow$ & 59.9 \\
12 & cop & -- & 49.7 & $\nearrow$ & 52.3 & -- & 51.8 & -- & 50.2 \\
13 & epist & -- & 43.9 & $\nearrow$ & 51.5 & $\nearrow$ & 56.6 & $\searrow$ & 56.2 \\
14 & fin & $\nearrow$ & 60.5 & -- & 52.9 & -- & 54.0 & $\searrow$ & 62.1 \\
15 & inf & -- & 47.7 & -- & 47.7 & $\searrow$ & 55.7 & $\searrow$ & 58.5 \\
16 & mark & -- & 58.3 & -- & 54.1 & -- & 50.7 & $\searrow$ & 57.9 \\
\rowcolor{lightgray}
17 & mean\_sent\_wc & $\nearrow$ & 59.5 & $\nearrow$ & 51.6 & $\searrow$ &  \textbf{68.0} & $\searrow$ &  \textbf{66.9} \\
18 & mhd & $\nearrow$ & 63.8 & -- & 42.4 & $\searrow$ &  \textbf{65.8} & $\searrow$ & 63.5 \\
19 & mpred & -- & 39.1 & $\nearrow$ & 54.4 & -- & -- & -- & 48.3 \\
20 & negs & -- & 41.6 & -- & 49.2 & $\searrow$ & 54.4 & -- & 44.0 \\
21 & nmod & $\nearrow$ & 61.0 & -- & -- & $\searrow$ &  \textbf{67.8} & $\searrow$ & 58.2 \\
\rowcolor{lightgray}
22 & nn & $\nearrow$ & 61.0 & $\searrow$ & 54.2 & $\searrow$ & 64.8 & $\searrow$ & 54.0 \\
23 & nnargs & $\nearrow$ & 55.6 & -- & 50.7 & $\searrow$ &  \textbf{67.4} & $\searrow$ & 55.8 \\
24 & nsubj & $\nearrow$ & 58.9 & -- & 46.9 & -- & 55.1 & $\searrow$ & 60.6 \\
25 & nummod & -- & 50.5 & -- & -- & $\nearrow$ & 44.7 & $\searrow$ & 54.1 \\
26 & obj & $\nearrow$ & 58.4 & -- & 46.5 & -- & 52.5 & $\searrow$ & \textbf{68.7} \\
\rowcolor{lightgray}
27 & obl & $\nearrow$ & 58.4 & $\nearrow$ & 54.5 & $\searrow$ &  \textbf{67.0} & $\searrow$ & 62.8 \\
28 & obl\_obj & -- & 50.1 & -- & 48.5 & -- & 50.1 & $\searrow$ & 59.6 \\
29 & parataxis & $\searrow$ & 55.9 & $\nearrow$ & 53.3 & -- & 50.0 & -- & -- \\
30 & pastv & $\nearrow$ & 63.0 & -- & 49.5 & $\nearrow$ & 60.6 & $\searrow$ & 64.0 \\
31 & ppron & -- & 41.2 & $\searrow$ & 52.5 & $\nearrow$ & 62.3 & -- & 52.8 \\
32 & prep & $\nearrow$ & 57.8 & -- & 48.7 & $\searrow$ & 64.4 & $\searrow$ & 54.0 \\
33 & relcl & $\nearrow$ & 52.1 & -- & 52.4 & -- & 47.0 & $\searrow$ & 62.9 \\
34 & ttr & $\nearrow$ & 60.3 & -- & 53.5 & $\nearrow$ & 50.0 & -- & 42.5 \\
35 & vo\_noun & -- & 50.4 & -- & 44.7 & -- & 50.4 & $\searrow$ & 60.9 \\
36 & vorfeld & $\searrow$ & 50.4 & -- & 52.1 & $\searrow$ & 55.1 & $\searrow$ & 54.0 \\
37 & xcomp & -- & -- & -- & 48.7 & $\nearrow$ & 51.7 & -- & 42.0 \\
\bottomrule
\end{tabular}%
		\captionof{table}{\label{tab:doc_svm_arrows}Document-level frequency comparison between targets and originals and macro F1 scores for single-feature classifiers. `--' indicates either no statistically significant difference ($p > 0.05$) between classes or cases where a single-feature classifier produced no predictions for one class in at least one fold, rendering the metric undefined. Arrows ($\nearrow$, $\searrow$) denote significantly higher or lower average feature frequency in target texts. Three features exhibiting statistically significant class differences for all directions and modes are highlighted. Univariate performances with F1 scores $> 65$ are in bold. Features are sorted alphabetically.}
\end{minipage}

% \vspace*{-1.0em}
\begin{minipage}{\linewidth}
\centering
\captionsetup{type=figure}
	% ---- Written ----
	\begin{subfigure}[t]{0.48\textwidth}
		\centering
		\includegraphics[width=\linewidth]{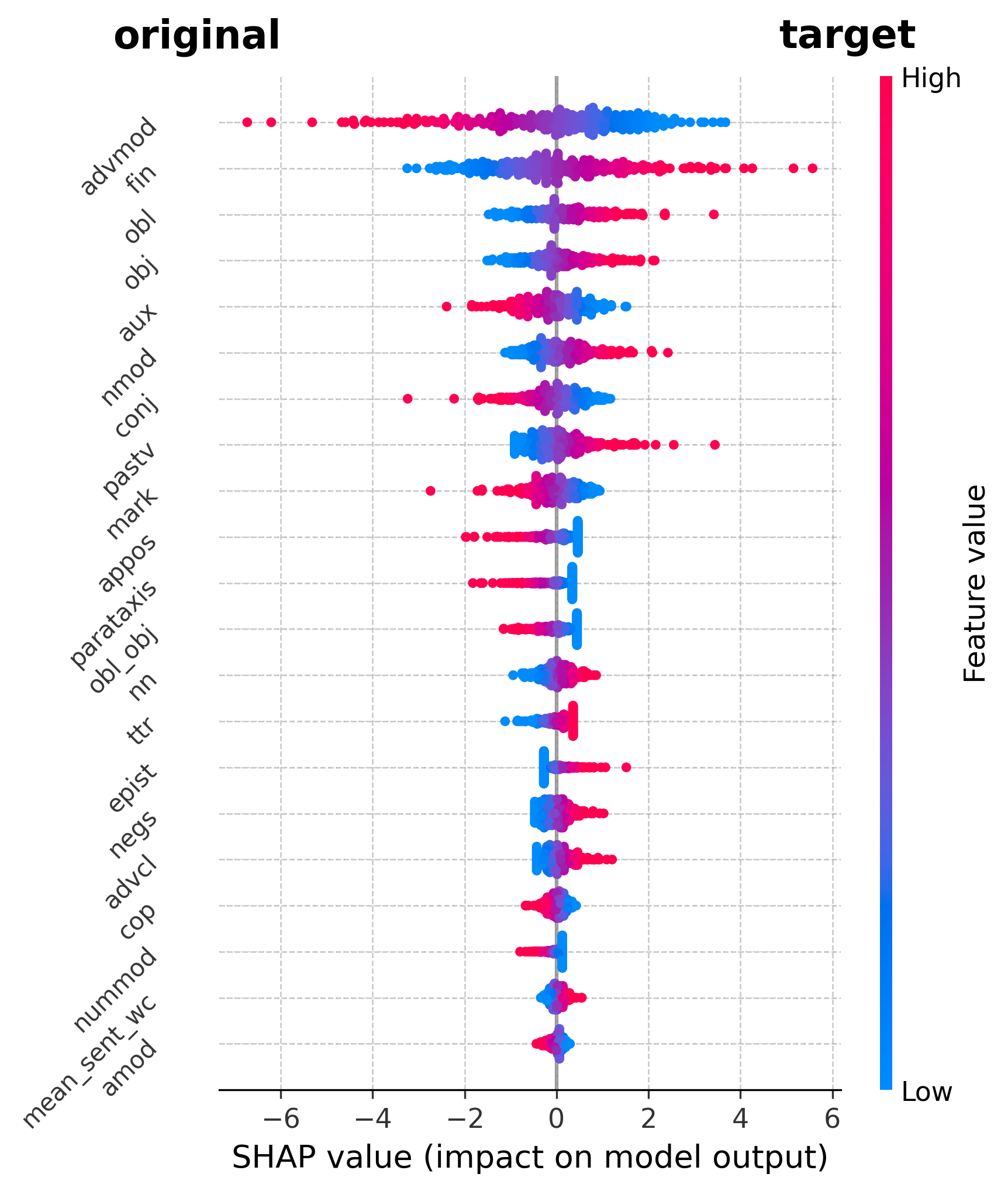}
		\caption{Written DE}
	\end{subfigure}
	\hfill
	\begin{subfigure}[t]{0.48\textwidth}
		\centering
		\includegraphics[width=\linewidth]{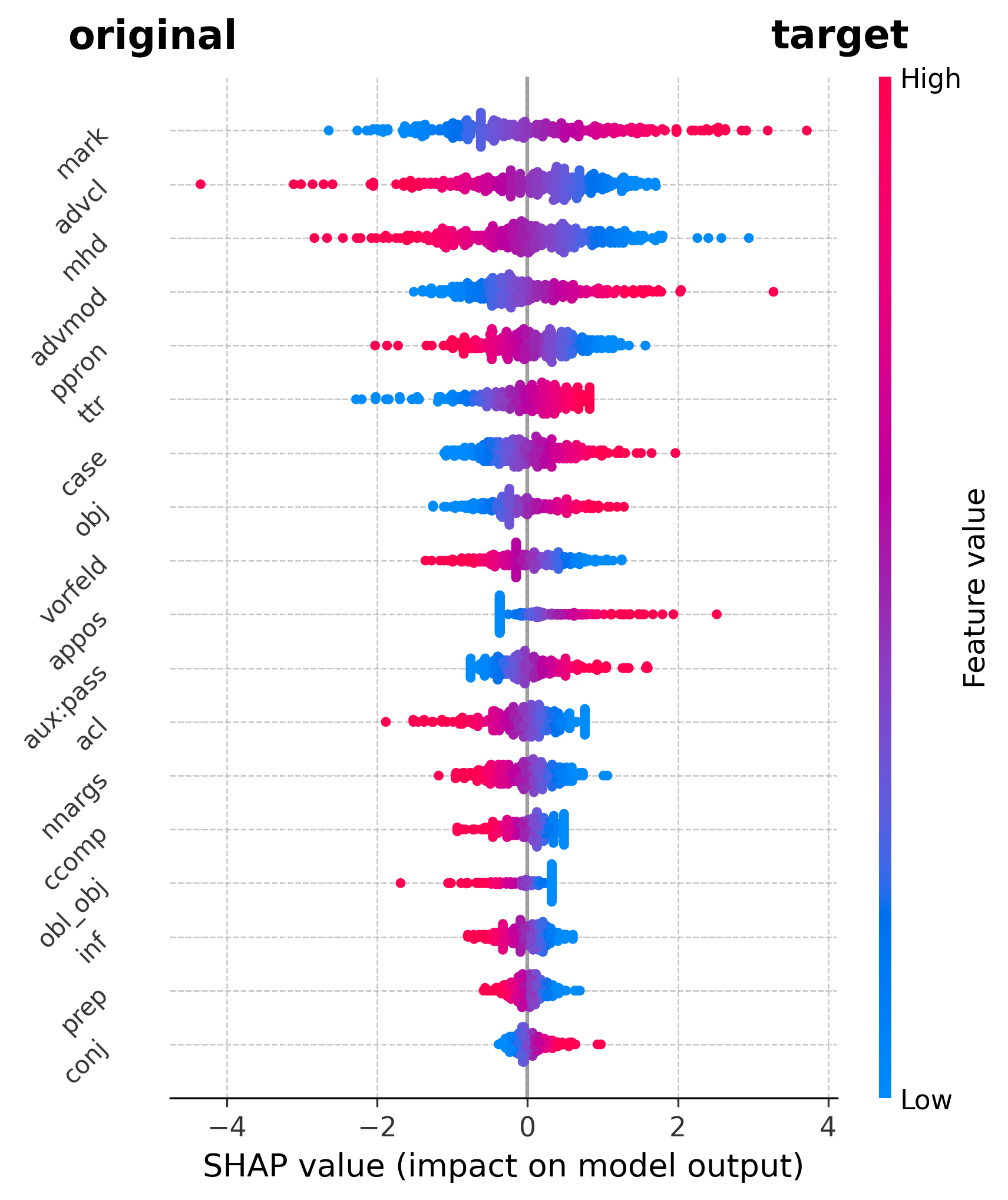}
		\caption{Written EN}
	\end{subfigure}

	% \vspace{1em}

	% ---- Spoken ----
	\begin{subfigure}[t]{0.48\textwidth}
		\centering
		\includegraphics[width=\linewidth]{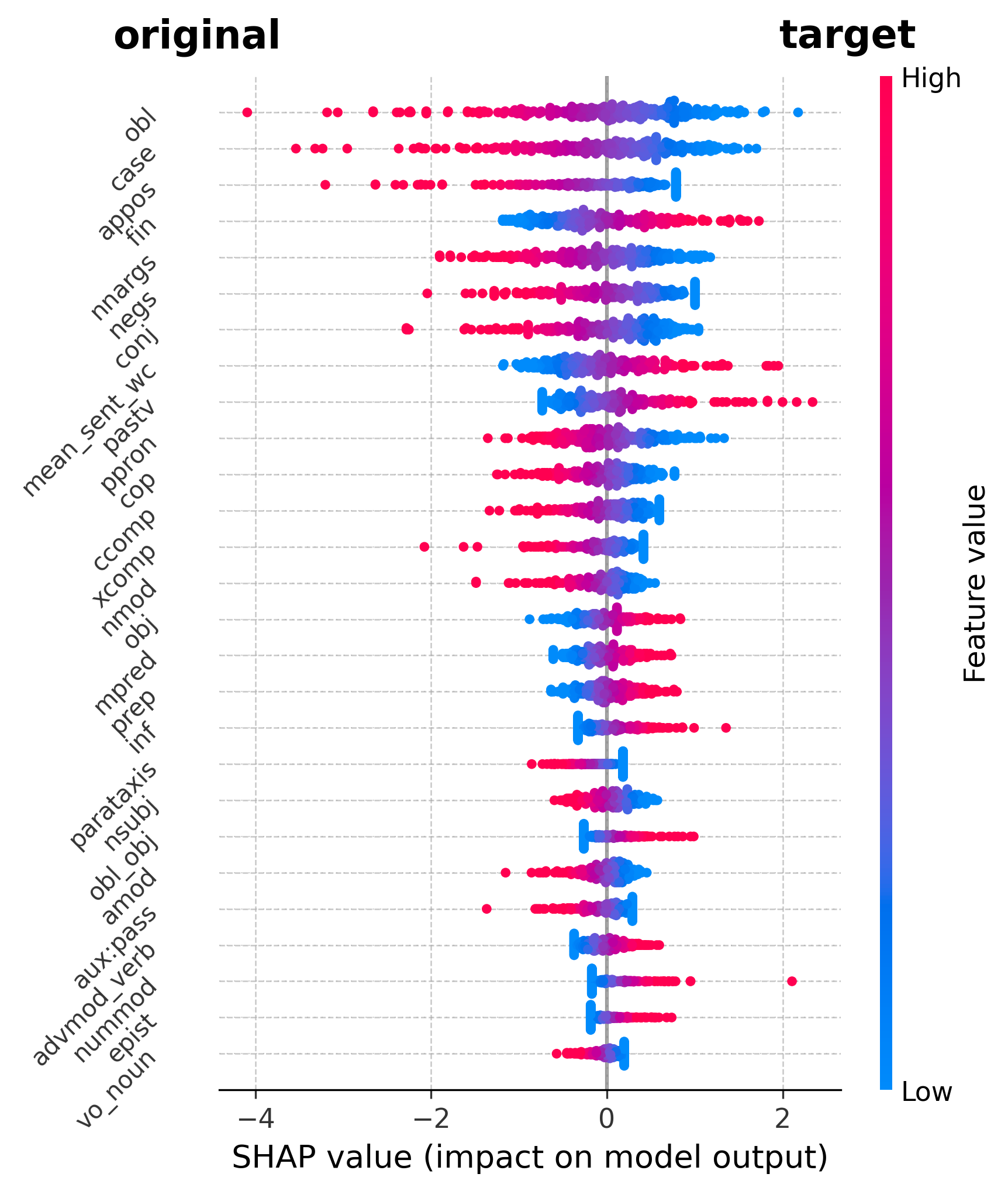}
		\caption{Spoken DE}
	\end{subfigure}
	\hfill
	\begin{subfigure}[t]{0.48\textwidth}
		\centering
		\includegraphics[width=\linewidth]{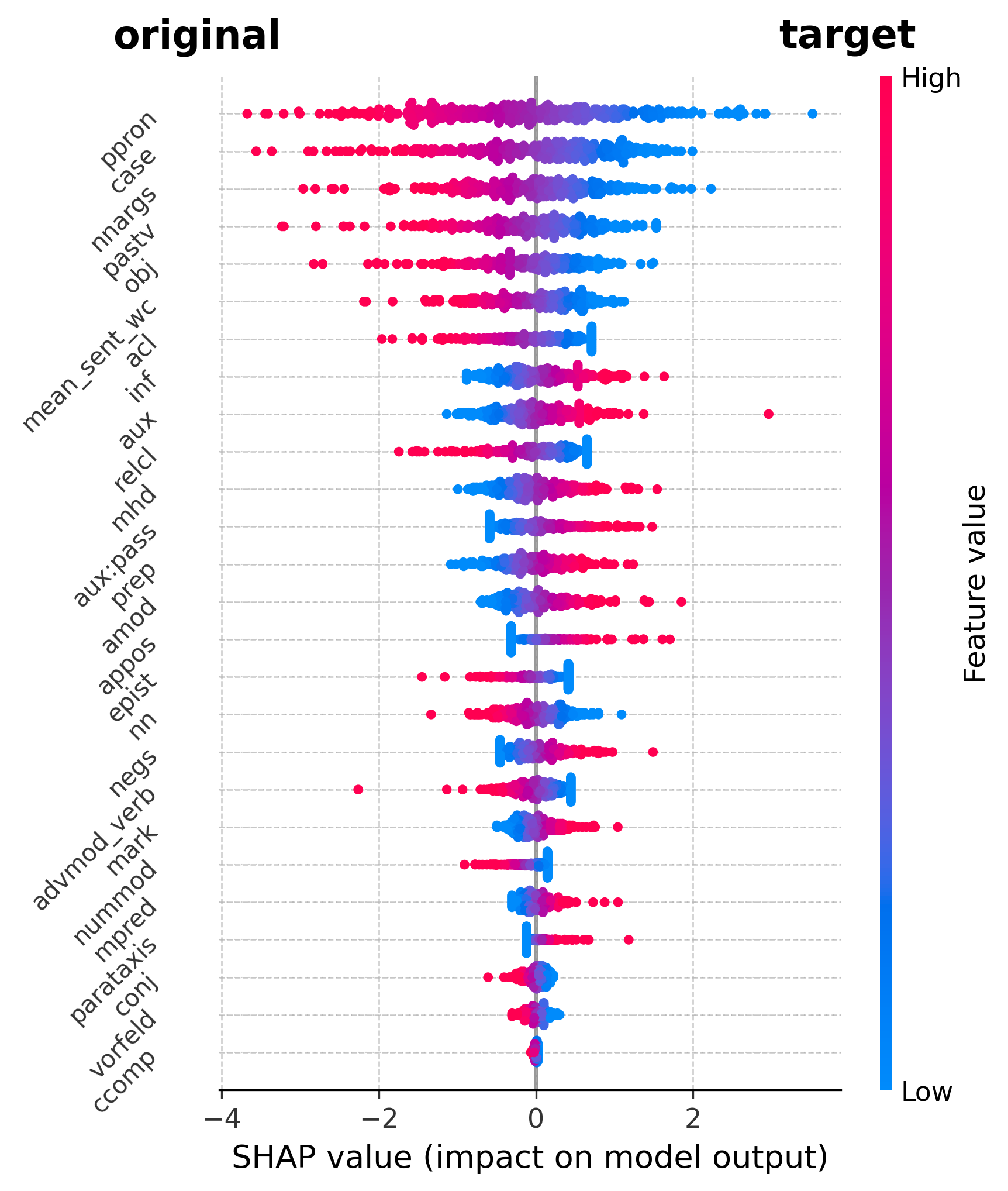}
		\caption{Spoken EN}
	\end{subfigure}
	\captionof{figure}{\label{fig:doc_svm_shap_beeswarm}SHAP beeswarm plots for document-level translationese classifications, showing \textit{selected features} contributions to class predictions, sorted by importance for each mode-language combination. Rows correspond to mediation mode (written vs.\ spoken), columns to target language (German vs.\ English). Each point corresponds to an individual segment; colour scale from red (high) to blue (low) encodes the feature value, while horizontal position reflects the SHAP value (in the log-odds scale), i.e.\ the impact of that feature on the model output.}
\end{minipage}
% \vfill

\clearpage
\twocolumn
\raggedbottom
\section{Appendix. Regression Details}\label{app:reg}
This addentix contains illustrations and supporting evidence for the claims in the main text with regard to regression analysis. 

\vspace{1em}
\begin{minipage}{\columnwidth}
\centering
	\captionsetup{type=figure} 
	\includegraphics[width=\columnwidth]{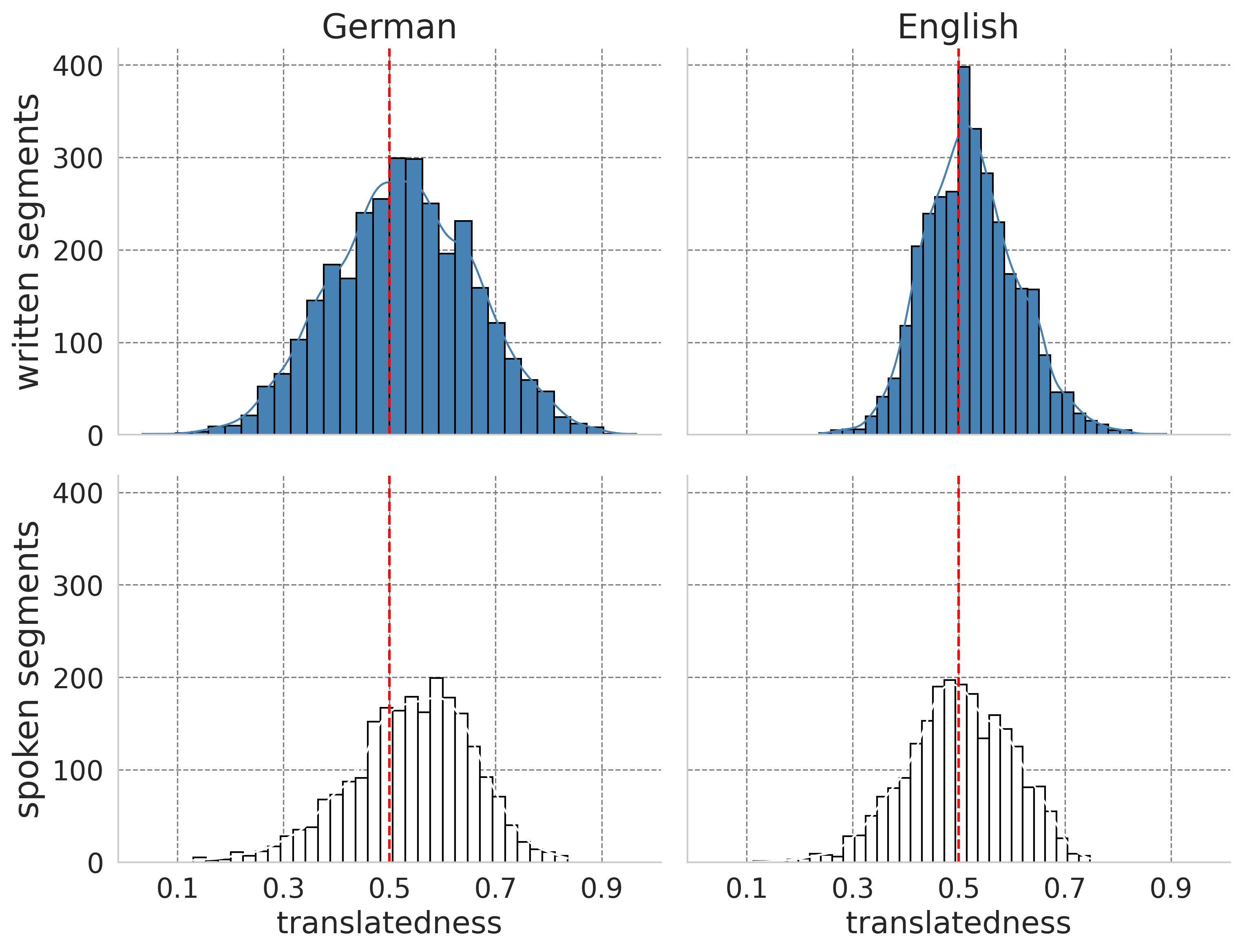}
	\caption{\label{fig:probs}Probability of being a translation (translatedness score) for target segments from the four segment-level translationese classifiers.}
\end{minipage}

\vspace{.5em}

\begin{minipage}{\columnwidth}
\captionsetup{type=table}
\resizebox{\columnwidth}{!}{%
\begin{tabular}{@{}lllrrrr@{}}
	\toprule
    % this is BASE results: 12 Feb 2026
	lpair & mode & approach & $\rho$ & $R^2$ & MAE     & selected/input \\ \midrule
	\multirow{6}{*}{deen}  & \multirow{3}{*}{written}   & source          & 0.27   & 0.06  & 0.07 & 8 / 11  \\  % 
	                        &                         & transfer        & 0.29   & 0.07  & 0.07 & 6 / 8  \\ % 
	                        &                         & source+transfer & \textbf{0.34}   & \textbf{0.10}  & 0.06  & 18 / 19  \\ \cmidrule(lr){2-7}% 
	                        & \multirow{3}{*}{spoken}   & source          & 0.25   & 0.06  & 0.08 & 6 / 12  \\ % 
	                        &                         & transfer        & 0.23   & 0.06 & 0.08 & 8 / 9  \\ % 
	                        &                         & source+transfer & \textbf{0.30}   & \textbf{0.10}  & 0.08 & 21 / 21  \\ \midrule % 
	
	\multirow{6}{*}{ende}  & \multirow{3}{*}{written}   & source          & 0.36   & 0.12  & 0.10 & 11 / 11  \\ % 
							&                         & transfer        & 0.44   & 0.18  & 0.09 & 8 / 8 \\ %  
	                        &                         & source+transfer & \textbf{0.47}   & \textbf{0.21}  & 0.09 & 19 / 19  \\ \cmidrule(lr){2-7} %             
	                        & \multirow{3}{*}{spoken}   & source          & 0.34   & 0.11  & 0.09 & 8 / 11 \\% 
	                        &                         & transfer        & 0.35   & 0.11  & 0.09 & 6 / 9 \\ %   
	                        &                         & source+transfer & \textbf{0.40}   & \textbf{0.16}  & 0.08 & 20 / 20  \\ % 
 		\bottomrule
\end{tabular}
}%
\caption{\label{tab:svr_diff} \textbf{Difficulty dimension-based analysis:} Results of linear SVM regression for features capturing source comprehension vs. transfer difficulty as well as combined models' results. Metrics include trend, fit, and error, alongside the count of features selected for the final model.}
\end{minipage}

\vspace{2em}
\begin{minipage}{\columnwidth}
\captionsetup{type=table}
\resizebox{\columnwidth}{!}{%
\begin{tabular}{@{}lllrrrr@{}}
\toprule
lpair                 & mode                     & approach     & $\rho$ & $R^2$ & MAE & selected/input \\ \midrule
\multirow{6}{*}{deen} & \multirow{2}{*}{written}                  & structure    & 0.28          & 0.06       & 0.07       & 11 / 13            \\
                      &  & IT           & 0.28          & 0.06       & 0.07       & 6 / 6              \\
                      \cmidrule(lr){2-7}
                      & \multirow{2}{*}{spoken}  & structure    & \textbf{0.28}          & \textbf{0.08}       & 0.08       & 6 / 13             \\
                      &                          & IT           & 0.22          & 0.05       & 0.08       & 7 / 8              \\
                      \midrule
\multirow{6}{*}{ende} & \multirow{2}{*}{written} & structure    & 0.37          & 0.13       & 0.10        & 12 / 13            \\
                      &                          & IT           & \textbf{0.42}          & \textbf{0.17}       & 0.10        & 6 / 6              \\
                      \cmidrule(lr){2-7}
                      & \multirow{2}{*}{spoken}  & structure    & \textbf{0.36}          & \textbf{0.13}       & 0.08       & 13 / 13            \\
                      &                          & IT           & 0.28          & 0.07       & 0.09       & 5 / 7              \\
                      \bottomrule 
\end{tabular}
}%
\caption{\label{tab:svr_feat_type} \textbf{Analysis by feature category:} Results of linear SVM regression for difficulty features grouped by their type (IT vs structural features). In addition to performance metrics, the number of RFECV-selected features, relative to the total post-filter input is reported. The combined models' performance is the same as in Table~\ref{tab:svr_diff}.}
\end{minipage}

\vspace{1em}

\begin{minipage}{\columnwidth}
	
	\centering  
    \captionsetup{type=figure}
	\includegraphics[width=\columnwidth]{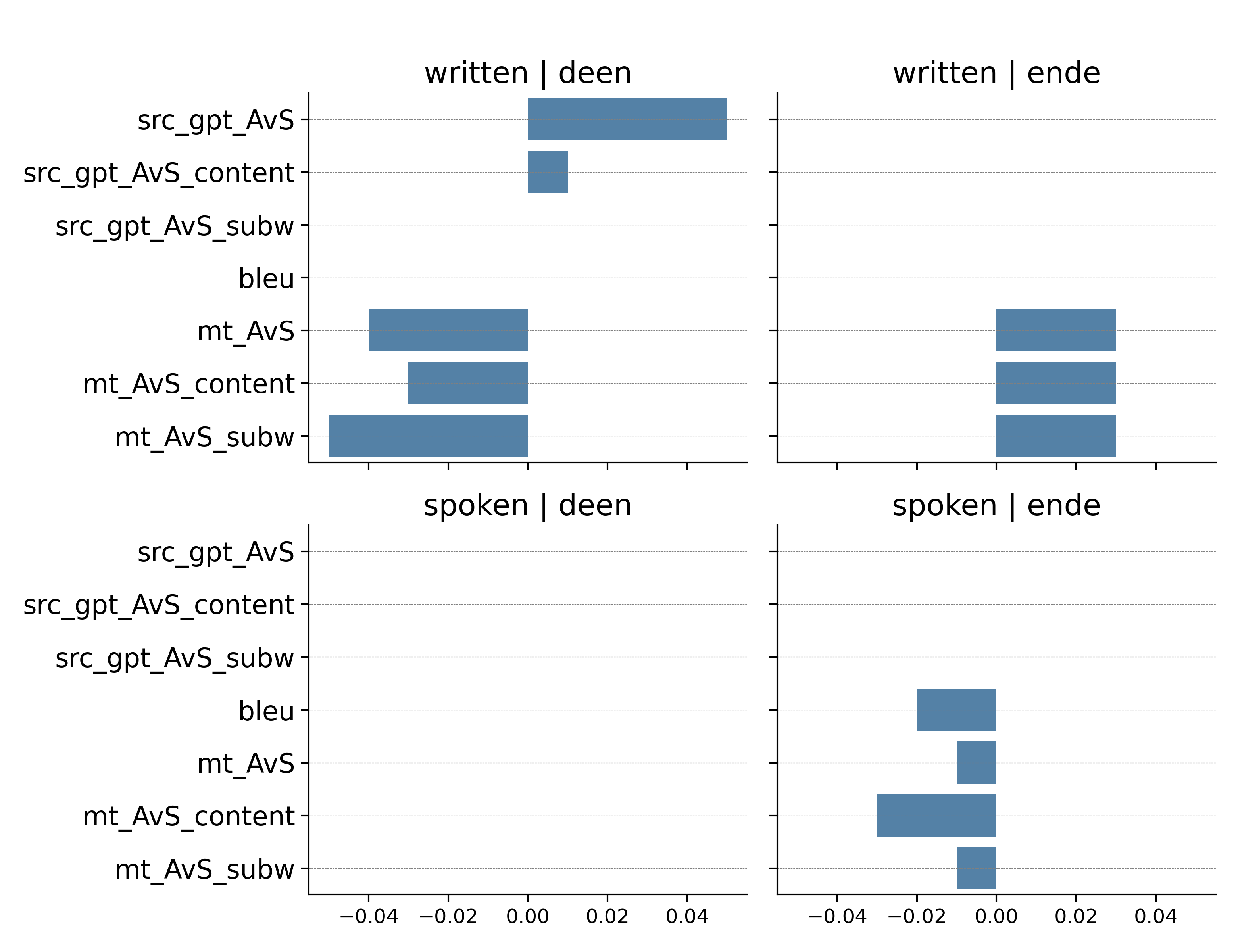}
	\caption{\label{fig:base_ft}LLM adaptation effect on the regression outcomes, calculated as direction-aligned $\delta$ $\rho$ (Spearman, ft - base). The bars represent how much the values of LLM-dependent features in Table~\ref{tab:svr_uni} (and potentially in Table~\ref{tab:svr_diff}) would increase or decrease, if we used fine-tuned models.}
\end{minipage}

\vspace{1em}
\begin{minipage}{\columnwidth}
	\centering 
    \captionsetup{type=figure}
	\includegraphics[width=\columnwidth]{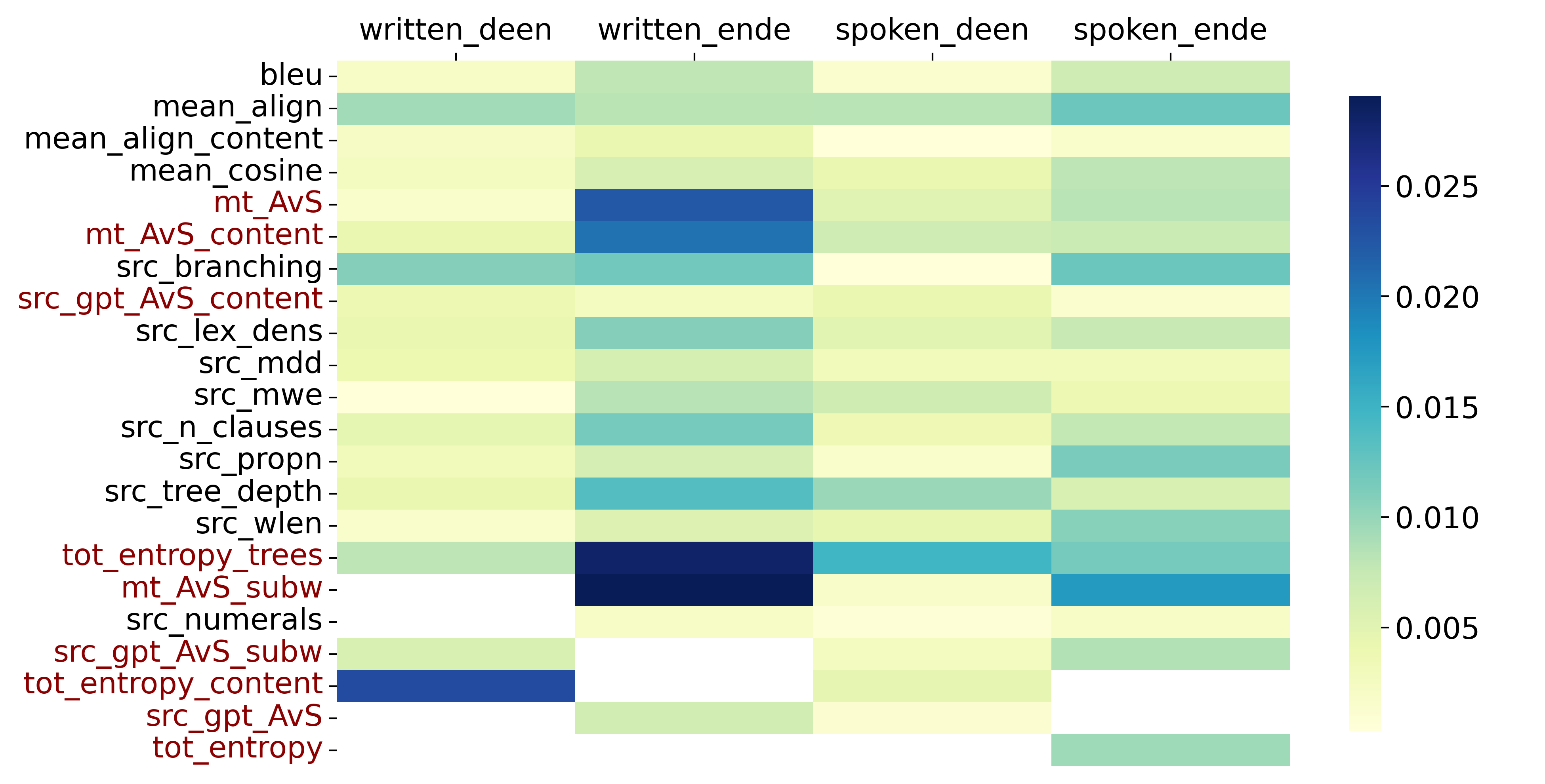}
	\caption{\label{fig:shared_seg_svr}Regression feature importances: Average absolute SHAP values for each translation task difficulty feature across mode-direction combinations. Darker shades indicate higher feature importance, blank cells indicate no contribution in that regression task. Information-theoretical indicators are in red. Features are ordered by the number of experiments in which they were selected, then alphabetically.}
\end{minipage}

\clearpage
\onecolumn
\begin{minipage}{\linewidth}
\centering
% {\renewcommand{\arraystretch}{0.9}
\captionsetup{type=table}

    \setlength{\tabcolsep}{0pt}
    \begin{tabular*}{\columnwidth}{@{\extracolsep{\fill}} ll | rcrc | rcrc @{}}
        \toprule
        \multirow{2}{*}{\#} & \multirow{2}{*}{Feature} & \multicolumn{2}{c}{written\_deen} & \multicolumn{2}{c}{written\_ende} & \multicolumn{2}{c}{spoken\_deen} & \multicolumn{2}{c}{spoken\_ende} \\ 
        \cmidrule(lr){3-4} \cmidrule(lr){5-6} \cmidrule(lr){7-8} \cmidrule(lr){9-10}
        & & mean & xy\_corr & mean & xy\_corr & mean & xy\_corr & mean & xy\_corr \\ 
        \midrule
        1  & bleu                  & 25.87 & --    & 21.71 & 0.10  & 14.47 & --    & 12.85 & -0.10 \\
        2  & mean\_align           & 0.84  & --    & 0.83  & 0.05  & 0.72  & --    & 0.72  & -0.13 \\
        3  & mean\_align\_cont.    & 0.83  & -0.04 & 0.83  & -0.05 & 0.70  & --    & 0.70  & -0.07 \\
        4  & mean\_cosine          & 0.80  & -0.04 & 0.79  & --    & 0.76  & 0.05  & 0.75  & -0.12 \\
        5  & mt\_AvS               & 10.99 & -0.06 & 12.59 & \textbf{0.23} & 12.02 & --    & 12.10 & -0.18 \\
        6  & mt\_AvS\_content      & 11.82 & -0.10 & 16.07 & \textbf{0.24} & 12.42 & --    & 15.11 & -0.13 \\
        7  & mt\_AvS\_subw         & 10.41 & -0.11 & 11.08 & \textbf{0.27} & 10.94 & -0.04 & 10.89 & -0.18 \\
        8  & src\_branching        & 1.63  & \textbf{0.23} & 2.27  & \textbf{0.24} & 1.55  & -0.15 & 1.89  & \textbf{-0.22} \\
        9  & src\_gpt\_AvS         & 6.57  & -0.10 & 6.26  & -0.04 & 7.34  & 0.10  & 6.74  & --    \\
        10 & src\_gpt\_AvS\_cont.  & 8.77  & -0.11 & 8.47  & --    & 9.19  & 0.07  & 8.26  & --    \\
        11 & src\_gpt\_AvS\_subw   & 5.91  & -0.08 & 6.00  & -0.07 & 6.60  & 0.09  & 6.36  & --    \\
        12 & src\_lex\_dens        & 0.41  & --    & 0.41  & --    & 0.42  & --    & 0.39  & -0.15 \\
        13 & src\_mdd              & 3.87  & 0.17  & 3.20  & 0.19  & 3.40  & -0.14 & 2.92  & -0.12 \\
        14 & src\_mwe              & 0.00  & 0.04  & 0.02  & 0.15  & 0.01  & -0.12 & 0.02  & -0.15 \\
        15 & src\_n\_clauses       & 2.03  & 0.06  & 2.40  & \textbf{0.25} & 1.82  & -0.13 & 2.08  & --    \\
        16 & src\_numerals         & 0.01  & --    & 0.01  & 0.08  & 0.00  & --    & 0.01  & --    \\
        17 & src\_propn            & 0.02  & 0.11  & 0.03  & 0.13  & 0.02  & -0.04 & 0.03  & -0.16 \\
        18 & src\_tree\_depth      & 5.24  & 0.10  & 5.93  & \textbf{0.33} & 4.59  & -0.19 & 5.18  & -0.16 \\
        19 & src\_wlen             & 5.92  & -0.08 & 4.76  & 0.10  & 5.85  & --    & 4.56  & \textbf{-0.23} \\
        20 & tot\_entropy          & 51.95 & \textbf{0.22} & 65.14 & \textbf{0.32} & 41.00 & -0.19 & 50.25 & -0.18 \\
        21 & tot\_ent\_content     & 22.99 & \textbf{0.23} & 33.57 & \textbf{0.30} & 18.40 & -0.17 & 25.31 & \textbf{-0.21} \\
        22 & tot\_ent\_trees       & 15.32 & 0.12  & 19.24 & \textbf{0.35} & 10.67 & \textbf{-0.20} & 12.33 & \textbf{-0.25} \\ 
        \bottomrule
        \end{tabular*}
            \caption{\label{tab:svr_uni} Results of univariate analysis of difficulty indicators, sorted alphabetically. \textit{mean} denotes the average feature value; \textit{xy\_corr} represents the Spearman correlation between the feature and the translatedness score. Correlations with an absolute value $> 0.20$ are bolded to highlight primary predictors. Dash (--) indicate non-significant correlations ($p > 0.05$).}

\end{minipage}

\begin{minipage}{\linewidth}
\centering
\captionsetup{type=figure}
	% ---- Written ----
	\begin{subfigure}[t]{0.48\textwidth}
		\centering
		\includegraphics[width=\linewidth]{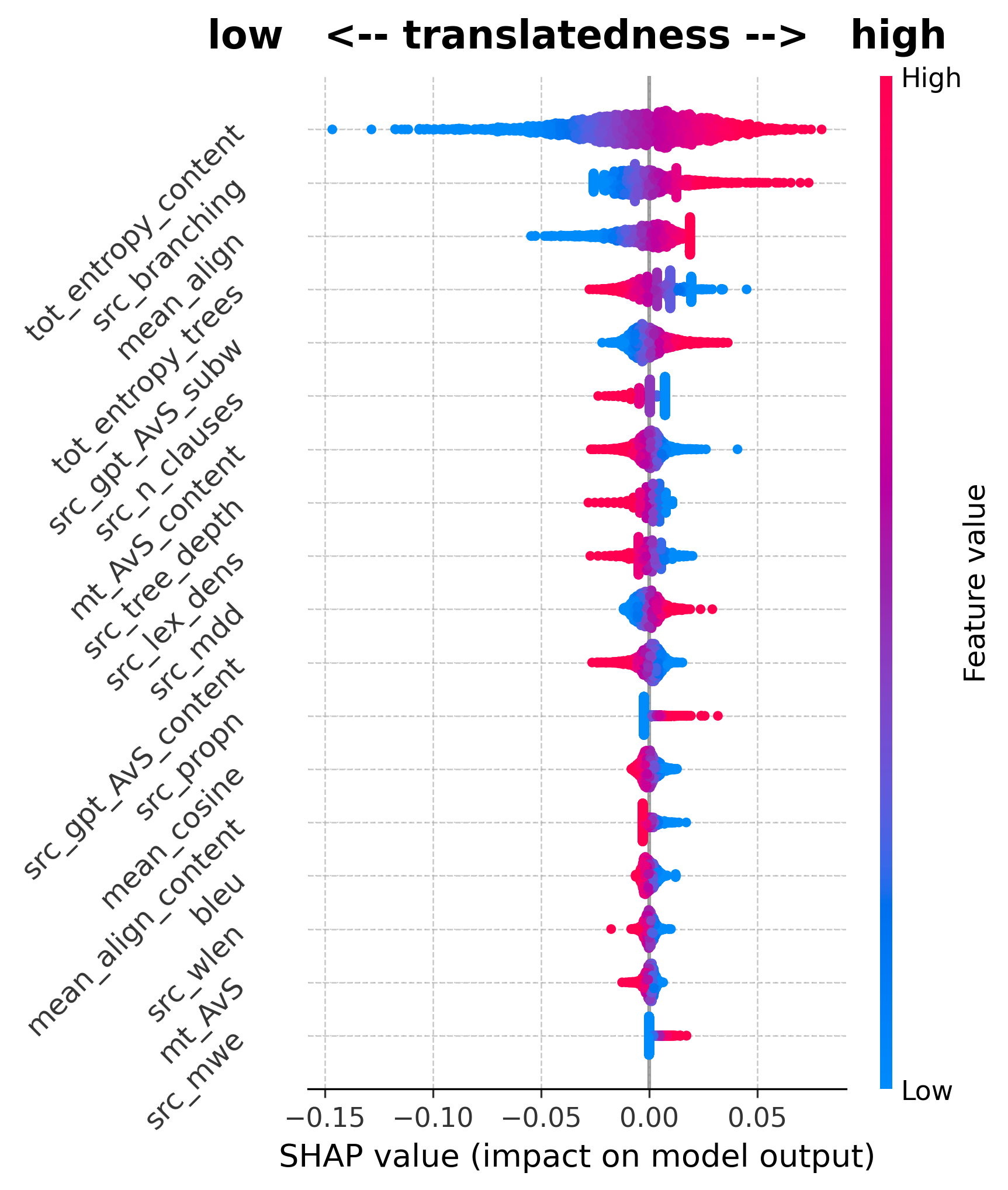}
		\caption{Written DE-EN}
	\end{subfigure}
	\hfill
	\begin{subfigure}[t]{0.48\textwidth}
		\centering
		\includegraphics[width=\linewidth]{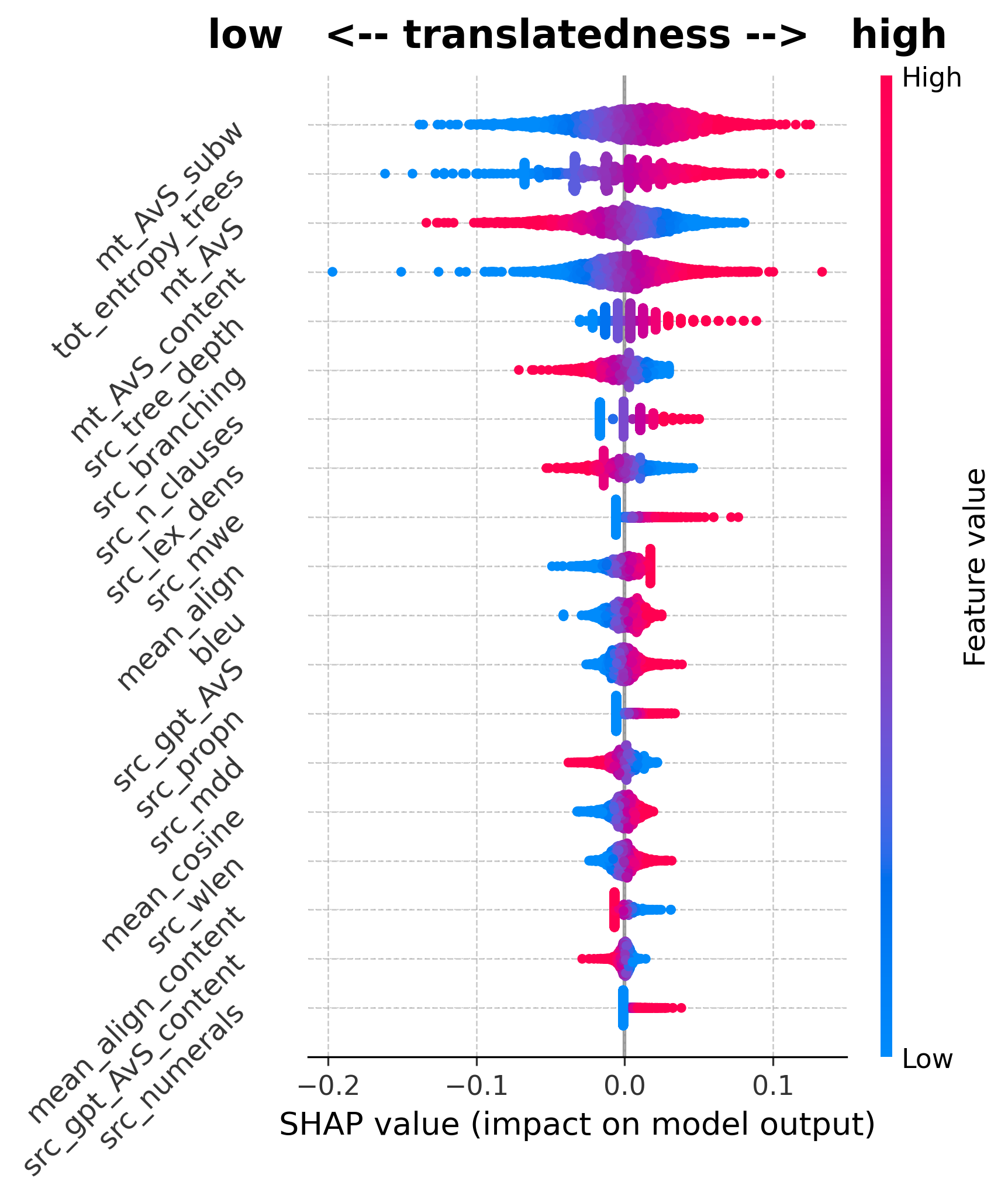}
		\caption{Written EN-DE}
	\end{subfigure}

	% \vspace{1em}

	% ---- Spoken ----
	\begin{subfigure}[t]{0.48\textwidth}
		\centering
		\includegraphics[width=\linewidth]{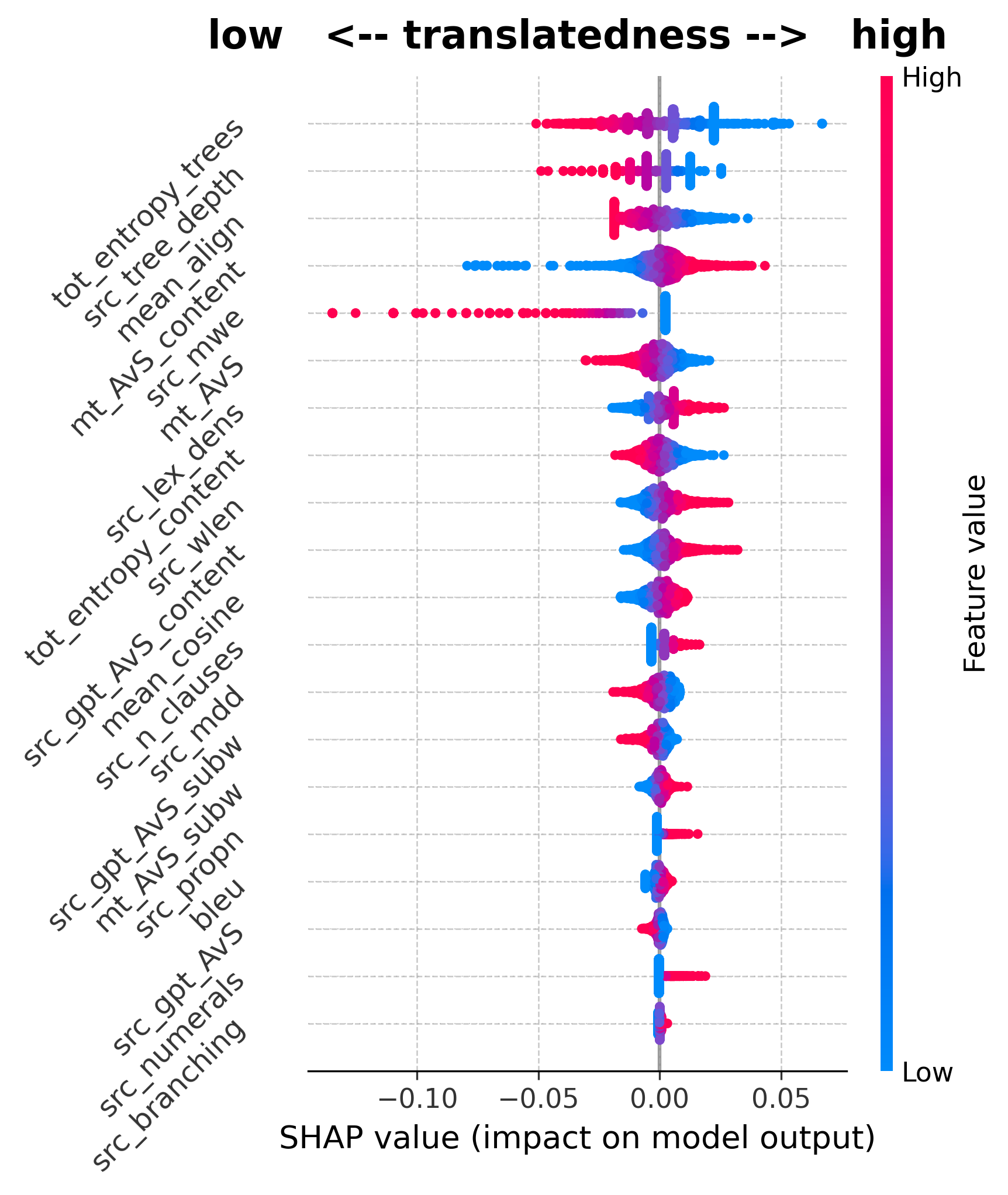}
		\caption{Spoken DE-EN}
	\end{subfigure}
	\hfill
	\begin{subfigure}[t]{0.48\textwidth}
		\centering
		\includegraphics[width=\linewidth]{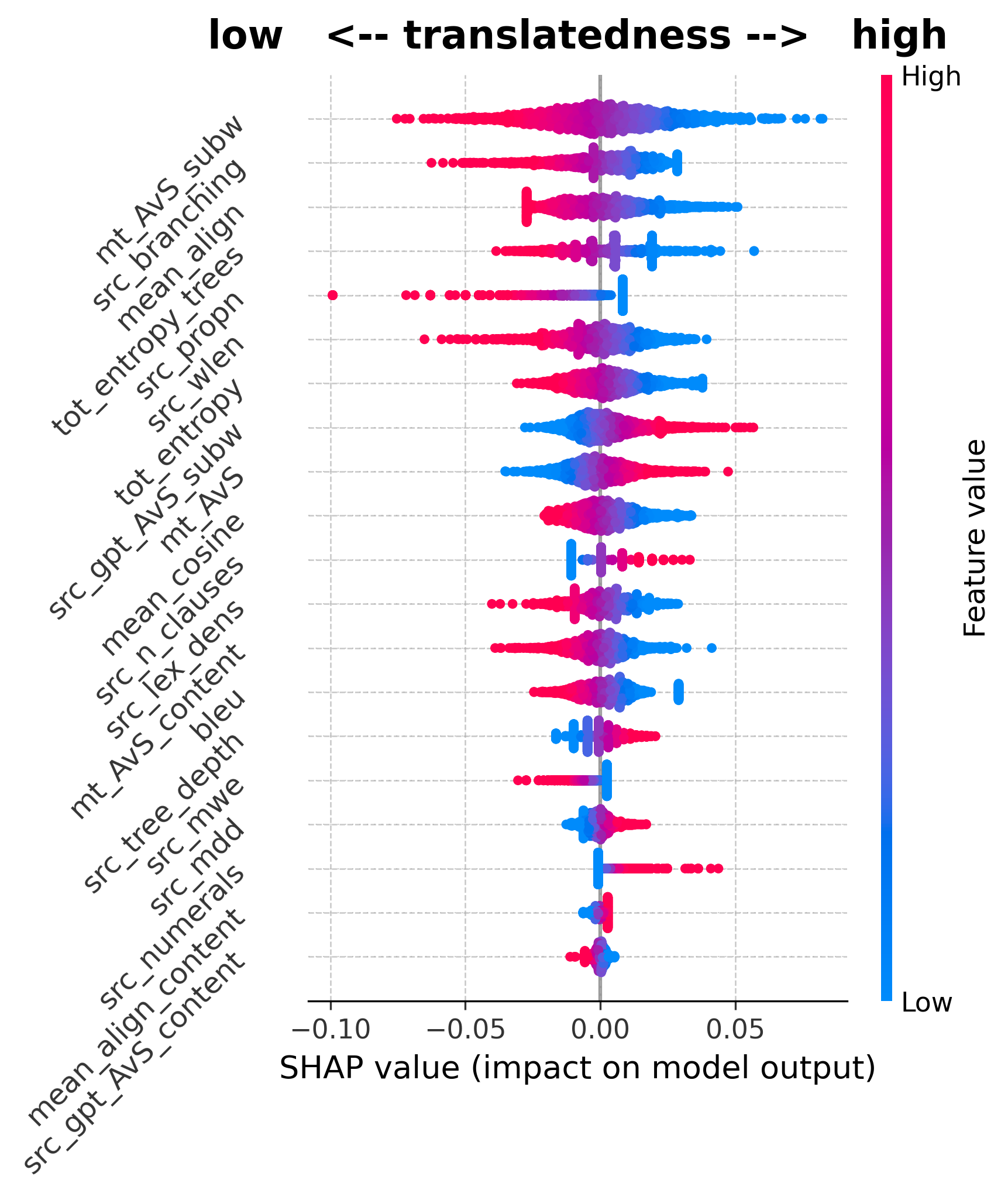}
		\caption{Spoken EN-DE}
	\end{subfigure}
	\captionof{figure}{\label{fig:svr_shap_beeswarm}SHAP beeswarm plots for regression, showing the impact of \textit{selected features} on the predicted translatedness scores. Features are sorted by mean absolute SHAP value, indicating their relative contribution to the deviation from the model's base value for each mode-language combination. Rows correspond to mediation mode (written vs.\ spoken), columns to target language (German vs.\ English). Each point corresponds to an individual segment; colour scale from red (high) to blue (low) encodes the feature value, while horizontal position reflects the SHAP value (in the log-odds scale), i.e.\ the impact of that feature on the model output.}
\end{minipage}

\end{document}